\documentclass{article}

\usepackage[preprint]{neurips_2026}
\usepackage[T1]{fontenc}
\usepackage{stfloats}
\usepackage{placeins}
\usepackage{xr}
\usepackage{xcolor}
\usepackage{amsthm}
\usepackage{verbatim}
\usepackage{booktabs}
\usepackage{multirow}
\usepackage{rotating}
\definecolor{improved}{RGB}{0,128,0}
\definecolor{degraded}{RGB}{192,0,0}
\usepackage{amsmath,amssymb}
\usepackage{graphicx}
\usepackage{capt-of}
\usepackage[hypertexnames=false]{hyperref}

\newcommand{\method}{AC}
\newcommand{\aclinear}{AC-Linear}
\newcommand{\acmlp}{AC-MLP}
\newcommand{\acridge}{AC-Density}
\newcommand{\tprlow}{TPR@FPR=$10^{-3}$}

\IfFileExists{preamble.tex}{\usepackage{times}
\usepackage{epsfig}
\usepackage{graphicx}
\usepackage{amsmath, amssymb}
\usepackage{booktabs}
\usepackage{multirow}
\usepackage{rotating}  
\usepackage{array}
\usepackage{float}
\usepackage{stfloats}
}{}

\title{Adaptive Calibration for Fair and Performant Facial Recognition}

\author{Ryan Brown \hfill Chris Russell\\
University of Oxford\\
{\tt\small firstname.lastname@oii.ox.ac.uk}
}

\begin{document}

\maketitle
\raggedbottom

\begin{abstract}
We introduce Adaptive Calibration (\method), a novel calibration strategy for facial recognition that maps cosine similarity between normalized embeddings to well-calibrated probabilities. By incorporating local context into calibration, Adaptive Calibration corrects for a fundamental mismatch in cosine similarity, whereby the same distance can correspond to different match probabilities in different embedding regions. Our approach improves both overall performance and results in a fairer calibration without requiring demographic metadata.

Our approach often improves worst-group ranking and probability quality across a variety of pretrained models and standard benchmarks, with gains in low-FPR verification performance depending on the setting. \method\  provides a practical solution for equitable facial recognition, without requiring demographic group annotations, and while improving overall performance. Our method provides continuous, region-specific calibration, and we examine per-group performance to assess ``leveling down'' where fairness comes at the cost of degraded performance for some groups.
\end{abstract}

\section{Introduction}

Face verification has been adopted across border control, device authentication, surveillance, and law enforcement \citep{sepas-moghaddam_face_2019, grother_face_2019, hill_facial_2024, kotwal_review_2025}. Modern systems map face images to high-dimensional embeddings and compare pairs by cosine similarity. Despite substantial accuracy progress, deployed systems still exhibit demographic disparities, with error rates that can be an order of magnitude higher for individuals with dark skin tones than for those with lighter skin \citep{grother_face_2019, buolamwini_gender_2018}.

\begin{figure*}[!t]
    \centering
    \IfFileExists{overview2.pdf}{\includegraphics[width=0.9\textwidth]{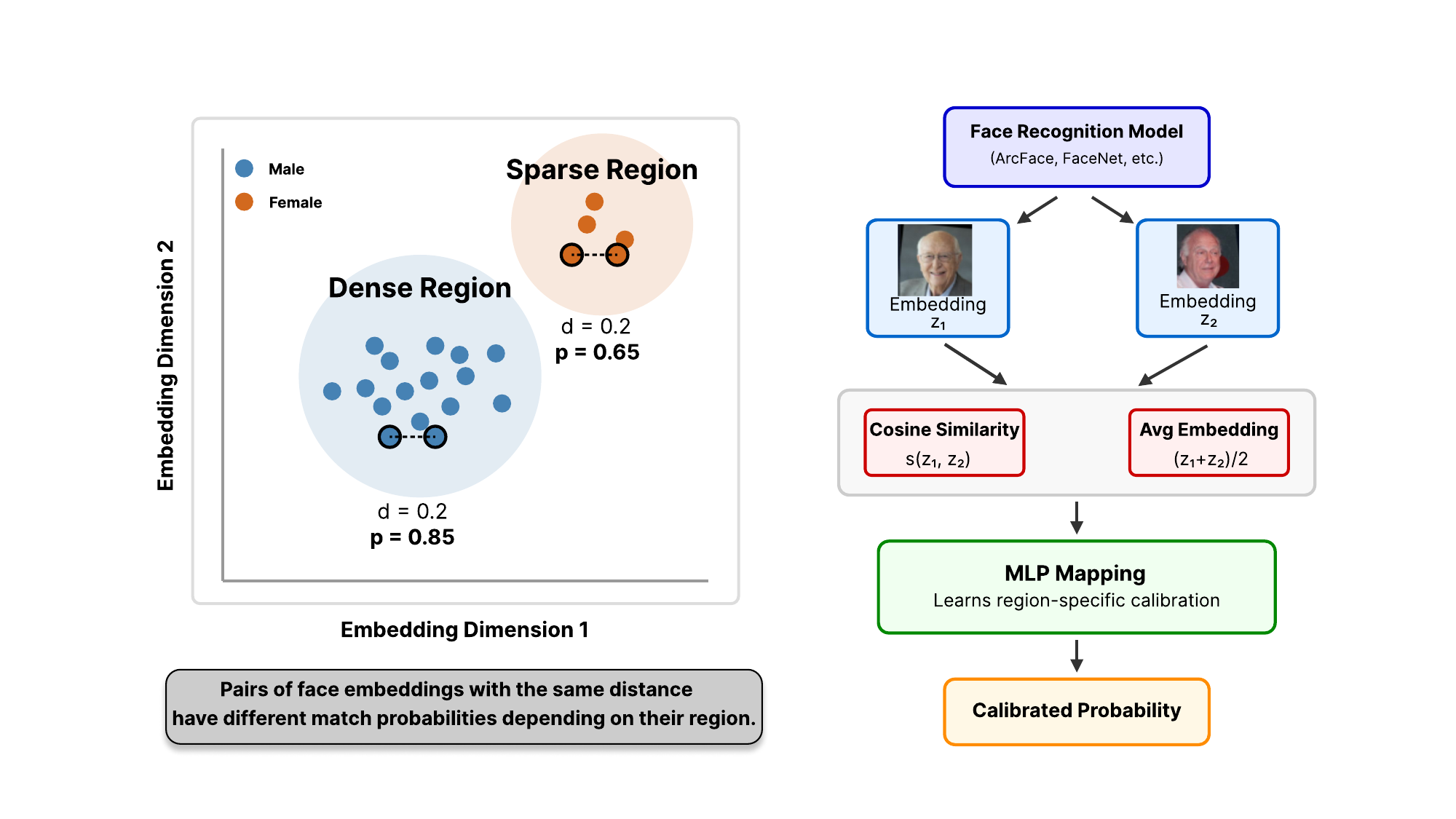}}{%
        \fbox{\parbox[c][1.5in][c]{0.85\linewidth}{\centering Method overview figure: \texttt{overview2.pdf}}}}
    \caption{Adaptive Calibration. Identical cosine distances in the embedding space correspond to different match probabilities in different regions of the space. \method{} uses cosine similarity together with the average pair location to learn a continuous, region-specific calibration through a logistic, MLP, or local-residual mapping.}
    \label{fig:ovi2}
\end{figure*}

Unlike classification, which is typically trained in a probabilistic framework \citep{prince2023understanding}, modern face recognition learns embeddings using distance-based objectives such as contrastive or angular-margin losses \citep{electronics15020338}. These models transform face images into high-dimensional embeddings, where the distance (cosine similarity) between normalized pairs is treated as an identity-similarity score. Although effective, the outputs of such systems do not map cleanly onto probabilities. Two pairs of faces with similar cosine distances can have different probabilities of being correct matches, depending on where they lie in the embedding space \citep{salvador_faircal_2022}.

When deployed, all matches within a threshold distance $T$ are treated as correct, and the resulting errors fall disproportionately on minority groups underrepresented in the training data \citep{grother_face_2019, cherepanova_deep_2022}. This is unsurprising: most large-scale face-recognition training sets are highly imbalanced \citep{wang_racial_2019, robinson_face_2020, buolamwini_gender_2018}. Underrepresented groups occupy lower-density regions of the embedding space where embedding behavior is less constrained \citep{cherepanova_deep_2022}, and a global threshold that works well for majority groups often performs poorly for them, yielding higher false rejection rates if the threshold is too strict, or higher false acceptance rates if it is too lenient \citep{huang_deep_2019}.

This is fundamentally a mismatch between the training objective\footnote{Typically a margin or triplet loss that ensures correct matches are closer than incorrect matches but makes no assumption about absolute distances.} and how such systems are operationalized\footnote{All matches with distances smaller than $T$ are treated as correct.}. To correct this mismatch, we convert distances into calibrated probabilities using a locally varying calibration function. A single strictly increasing score-only mapping, as in standard monotonic calibration, preserves the ordering and threshold-based disparities of the original score \citep{rahimi_post-hoc_2020, pleiss_fairness_2017, ding_local_2021, salvador_faircal_2022}.

We introduce Adaptive Calibration (\method), an approach that learns a continuous, locally varying calibration function as illustrated in Figure~\ref{fig:ovi2}. Operating as a post-hoc addition to any pretrained face-recognition system, \method{} achieves superior fairness--accuracy trade-offs and frequently outperforms existing approaches on operational metrics across modern recognition backbones (Tables~\ref{tab:tpr-by-backbone},~\ref{tab:mgauroc-by-backbone}, and~\ref{tab:ld-by-backbone}). Unlike existing locally varying approaches such as FairCal \citep{salvador_faircal_2022} or FRAPP\'E \citep{tifrea_frappe_2024}, \method{} avoids ad-hoc clustering heuristics and is a direct application of a range of calibrated machine-learning methods. Linear variants can be integrated into existing verification pipelines with minimal custom code.

\paragraph{Contributions.}
\begin{itemize}
    
    \item We propose Adaptive Calibration (\method; Figure~\ref{fig:ovi2}), a method that maps local embedding context and similarity scores to calibrated match probabilities, addressing shortcomings that harm both accuracy and fairness. 

    \item Our method improves fairness and accuracy over existing post-hoc calibration techniques in many of the evaluated settings.

    \item We release the methods, evaluation criteria, and implementation details in the supplementary material to support fairer and more accurate deployment of facial recognition systems.

    \item We explain how predicting conditional match probabilities can yield calibration for groups recoverable from the calibrator inputs, and evaluate several demographic and attribute-defined groups.

\end{itemize}

\section{Related Work}

\paragraph{Facial recognition.}
Face-recognition systems leverage deep neural networks to transform raw images into discriminative embeddings, enabling identity verification \citep{electronics15020338}. Cosine similarity is used as the matching score \citep{deng_arcface_2022}, and a threshold on this score determines whether two face images belong to the same identity. The threshold is calibrated on validation data to achieve a desired trade-off between false positives and false negatives. Modern face-recognition models predominantly rely on deep convolutional networks trained with angular-margin losses, including SphereFace \citep{liu_sphereface_2018}, CosFace \citep{wang_cosface_2018}, GhostFaceNets \citep{alansari_ghostfacenets_2023}, ArcFace \citep{deng_arcface_2022}, AdaFace, and MagFace. These methods enforce a hyperspherical embedding space with clear angular margins between identities, but the resulting score is not a calibrated probability.

\paragraph{Fairness in face recognition.}
Unlike fair classification, where a range of conflicting definitions have been proposed and can be reliably enforced for tabular data\cite{weerts_fairlearn_2023} and standard vision/NLP tasks\cite{delaney_oxonfair_2024}, fair facial recognition has settled on common fairness measures based around calibration. Of particular importance is the notion of \emph{group calibration}\footnote{Here we use a variant of group calibration that allows for any distance measure between a pair of variables, rather than a probability score computed for one variable.}, which says that for a pair of images at distance $T$, the probability that they share an identity should not depend on which protected groups they belong to:
\[
P\bigl(I_x = I_y \mid d(x,y)=T,\,x,y\in A_1\bigr)
=
P\bigl(I_x = I_y \mid d(x,y)=T,\,x,y\in A_2\bigr),\quad \forall A_1,A_2\in \mathcal{A}.
\]
Here $I_x$ and $I_y$ are the identities of images $x$ and $y$, $d(x,y)$ is the cosine distance between them, and $\mathcal{A}$ is the set of protected groups.

When operationalizing facial recognition, all matches closer than a threshold $T$ are accepted, so a related desideratum is
\[
P(I_x = I_y\mid d(x,y)\leq T,\, x,y\in A_1)
=
P(I_x = I_y\mid d(x,y)\leq T,\, x,y\in A_2),\quad \forall A_1,A_2\in \mathcal{A}.
\]
This corresponds to the notion of \emph{equal precision}, i.e., the probability of an accepted match being correct is the same for all groups. Pointwise group calibration does not generally imply equal precision after thresholding: accepted-score distributions can differ between groups. Equal score distributions are a sufficient additional condition for this implication.

\paragraph{Beyond disparity-based fairness measures.}
Any performance measure can be turned into a fairness measure by computing the measure per group and reporting the disparity. Minimizing these disparities directly incentivizes ``leveling down'' \citep{mittelstadt_unfairness_2023}, where performance for the best group is degraded to equalize outcomes. A more robust strategy is to report the measure on the worst-performing group, which prioritizes the worst-served group and answers the deployment-relevant question ``is this system good enough for all protected groups?''. 
These observations motivate methods that improve overall calibration and accuracy, particularly for the groups where the system performs worst. We therefore report global TPR at a low-FPR operating point as the headline accuracy metric, and use worst per-group AUROC together with worst-group Brier score as the headline fairness metrics.

\paragraph{Post-processing fairness for face verification.}
Post-processing methods adjust a trained model's outputs to reduce bias while seeking to maintain accuracy \citep{salvador_faircal_2022, kotwal_review_2025, dhar_pass_2021, linghu_score_2024, dhar_towards_2020, terhorst_post-comparison_2020, conti_mitigating_2024}. Some directly produce calibrated probabilities, while others modify distances and require an additional calibration step. \emph{FairCal} \citep{salvador_faircal_2022} applies $K$-means clustering on calibration embeddings to form pseudo-demographic groups, then applies beta calibration within each cluster. It does not require sensitive attributes at inference, but its effectiveness depends on the choice of $K$ and the discreteness of the cluster partition. \emph{AGENDA} \citep{dhar_towards_2020} and \emph{PASS} \citep{dhar_pass_2021} use adversarial training to suppress demographic information in embeddings; they require sensitive labels for training and can reduce recognition accuracy. \emph{FTC} \citep{terhorst_comparison-level_2020} learns a fair similarity classifier directly, with a fairness penalty that can also reduce accuracy. \emph{GST} \citep{robinson_face_2020} assigns separate decision thresholds per group; while effective, it requires group membership at inference. \emph{FSN} \citep{terhorst_post-comparison_2020} normalizes scores per cluster to equalize FPR at a fixed operating point, and typically requires beta calibration to produce probabilities. \emph{Oracle} calibration \citep{salvador_faircal_2022} fits a per-group calibrator using true demographics and is treated as an upper bound. \emph{DemoNorm} \citep{linghu_score_2024} applies group-specific score normalization and similarly relies on demographic information at inference. \emph{FRAPP\'E} \citep{tifrea_frappe_2024} learns an additive correction with a fairness criterion; it is model-agnostic but requires a fairness-aware training phase with sensitive attributes. \emph{FALCON} \citep{al-refai_falcon_2025} is another label-free post-processing method, using local feature normalization in the embedding space to improve robustness across groups; it operates by normalizing features/scores.

\method{} differs from cluster-based and demographic-aware approaches in that it learns a continuous calibration function over score and embedding context, requires no demographic labels at inference, and does not rely on a hard partition of the embedding space.

\section{Adaptive Calibration}

We seek a calibrated face-verification system that uses no group annotations at training or inference time. The population motivation is simple: if $X$ denotes the calibrator inputs and $p(X)=\mathbb{E}[Y\mid X]$, then for any group $G$ recoverable from $X$, the tower property gives $\mathbb{E}[Y\mid p(X),G]=p(X)$. Thus, conditional-probability estimation can support calibration beyond explicitly labeled demographic axes. This is related to group-optimality results under recoverability assumptions \citep{singh_when_2023}.

This argument does not guarantee group calibration for a fitted finite-sample model. Recoverability must hold for the actual calibrator inputs, and may be lost when compressing two embeddings into their pair center and similarity. Model capacity, regularization, estimation error, and distribution shift also matter. To limit overfitting, we use a symmetric pair-center representation and consider three variants: a non-linear MLP; a linear logistic model; and a residual model with a local-density feature.

\subsection{Pair representation}

Let $z_1,z_2\in\mathbb{R}^d$ be normalized embeddings from a frozen backbone, with cosine similarity $s=z_1^\top z_2$. Define the pair midpoint $m=(z_1+z_2)/2$ and its normalized direction $\bar z=m/\|m\|_2$ when $m\ne0$. The implementations use $m$ for \aclinear{} and \acridge{}, and $\bar z$ for \acmlp{}. Both are symmetric in the two images and provide a continuous context signal away from the zero midpoint. Only $\bar z$ lies on the unit hypersphere; the midpoint norm satisfies $\|m\|_2=\sqrt{(1+s)/2}$, so these representations are not interchangeable for regularized linear models.

\subsection{MLP-based calibration (\acmlp)}

We employ a lightweight MLP $g_\phi$ that maps the pair representation to a calibrated probability, $p=g_\phi(\bar z,s(z_1,z_2))$. The input has $d+1$ dimensions: the normalized pair center concatenated with the cosine similarity. We use a single hidden ReLU layer with $1024$ units followed by a sigmoid output, and train with Adam for 5 epochs at learning rate $10^{-3}$ with batch size $512$ using binary cross-entropy, $\mathcal{L}_{\text{BCE}}=-N^{-1}\sum_{i=1}^{N}[y_i\log(p_i)+(1-y_i)\log(1-p_i)]$.

\subsection{Linear calibration (\aclinear)}
\label{ac:linear}
We train a logistic regression model on the concatenation of $m$ and the cosine similarity, $p=\sigma(w^\top[m,s(z_1,z_2)]+b)$, where $\sigma$ is the sigmoid, $w$ is a learned weight vector, and $b$ is a bias term. For fixed context $m$, the sign of the score coefficient determines whether $p$ increases or decreases with $s$; the implementation does not constrain this sign. Across different pairs, contextual corrections can change the ranking. The model is trained by minimizing binary cross-entropy on the calibration training folds.

\subsection{Local-density residual calibration (\acridge)}\label{sec:method-acridge}
\acridge{} adds a regularized, context-dependent correction to a score-only probability predictor. We first fit a Platt base classifier $p_{\text{base}}(s)=\sigma(as+b)$ on the training pairs.

\paragraph{Local-density measure.}
For a query midpoint $m$, the feature $\rho(m)=k^{-1}\sum_{n\in\mathrm{NN}_k(m)}\|m-m_n^{\text{train}}\|_2$ is the mean Euclidean distance to the $k=20$ nearest reference training midpoints. Larger values indicate sparser regions. For BFW, we use a reference subsample of $6{,}000$ training midpoints with a fixed random seed. Training queries include the query itself when it belongs to the reference set; held-out queries do not.

\paragraph{Ridge residual.}
We form $\phi(z_1,z_2)=[m,\rho(m),s(z_1,z_2)]\in\mathbb{R}^{d+2}$
and standardize each feature to zero mean and unit variance using statistics estimated on the training fold. 
We then fit a closed-form ridge regression $\Delta_\psi(\phi)=\psi^\top\phi+\psi_0$ to the probability residual of the base classifier, with targets $r_n=y_n-p_\text{base}(s_n)$ and regularization strength $\alpha_\text{ridge}=1.0$. Including both $m$ and $s$ in the residual lets the correction be score-dependent and location-dependent rather than a global density-aware Platt adjustment. The final probability is $p=\mathrm{clip}(p_{\text{base}}(s)+\Delta_\psi(\phi),0,1)$.

Squared loss fits the real-valued residuals $r_n\in[-1,1]$, while the ridge penalty shrinks coefficients toward zero. Ridge does not induce sparse coefficients or guarantee reversion to the base prediction in regions unsupported by calibration data.

All variants are trained per fold on the same calibration split as the competing methods. \aclinear{} uses $L_2$-regularized logistic regression with the implementation's default inverse regularization strength $C=1.0$ and LBFGS solver. The \acmlp{} architecture above has $527{,}361$ parameters for $d=512$; its capacity can be substantial relative to small calibration sets. \acridge{} additionally requires a nearest-neighbor query at inference. Appendix~\ref{app:checks} reports component and calibration-size analyses.

\paragraph{Calibration and re-ranking.}
A strictly increasing global score-only transformation preserves AUROC, including within groups. Context-dependent predictions can instead re-rank verification pairs, so AUROC gains measure improved discrimination rather than calibration alone. We evaluate probability quality separately using Brier score and reliability metrics. Related local calibration methods already consider local errors and fairness \citep{luo2022local}; parameterized and sample-dependent temperature scaling also use input-dependent corrections \citep{tomani2022parameterized,joy2023sample}. Input dependence alone does not guarantee preservation of rankings across examples. Our focus is the simple pair-context construction and its evaluation for face verification.

\section{Experimental Details}\label{sec:exp_setup}

\begin{figure*}[!t]
    \centering
    \IfFileExists{dv_rfw_fn_web.pdf}{\includegraphics[width=0.95\textwidth]{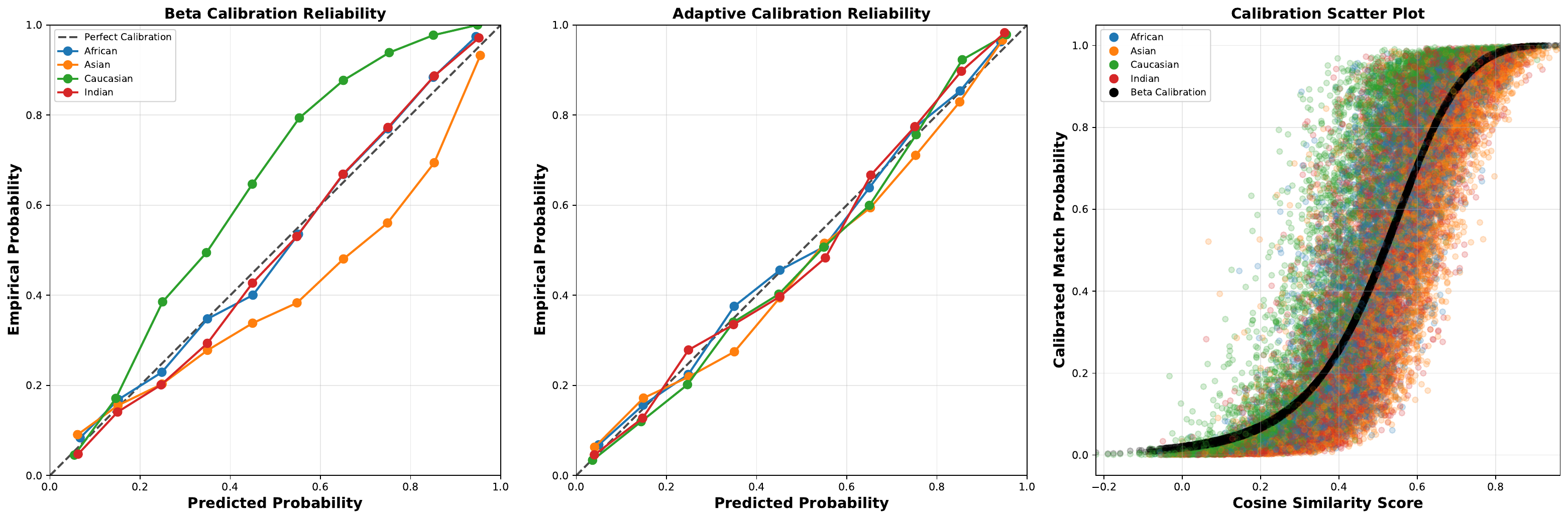}}{%
        \fbox{\parbox[c][1.5in][c]{0.9\linewidth}{\centering Reliability/score visualization: \texttt{dv\_rfw\_fn\_web.pdf}}}}
    \caption{Adaptive Calibration vs.\ beta calibration on RFW with FaceNet. Left: per-ethnicity reliability for beta calibration (closer to the diagonal is better). Center: per-ethnicity reliability for \method. Right: cosine similarity vs.\ match probability for both methods. Beta calibration (and standard thresholding) overestimates the probability of an Asian match being correct and underestimates the probability of a Caucasian match being correct, which substantially reduces the probability of a Caucasian false positive relative to Asian at any threshold. The wide overlap of per-group cosine values shows that a per-group adjustment is insufficient.}
    \label{fig:dv_rfw_fn_web}
\end{figure*}

For the main benchmark experiments, we follow a leave-one-fold-out cross-validation protocol analogous to FairCal \citep{salvador_faircal_2022}: calibration parameters are learned exclusively on training folds and applied to held-out test folds. We evaluate on three verification benchmarks. \textbf{RFW} \citep{wang_racial_2019} provides face pairs partitioned into four ethnic groups (African, Asian, Caucasian, Indian), with 10 folds covering 24,000 pairs. \textbf{BFW} \citep{robinson_face_2020} provides a larger benchmark with ethnicity and gender annotations and a 5-fold protocol over hundreds of thousands of pairs. \textbf{LFW} \citep{LFWTech} is  a standard public verification benchmark which is combined with FairFace-derived group annotations \citep{karkkainen_fairface_2021} for group analyses. It consists of 10 folds covering 6,000 pairs.

\paragraph{Backbones.}
We evaluate five modern face-recognition backbones as fixed feature extractors: FaceNet \citep{schroff_facenet_2015,sandberg_facenet_2018}, ArcFace R50 and ArcFace R100 from InsightFace \citep{deng_arcface_2022,guo_insightface_2025}, AdaFace, and MagFace. The main paper tables focus on ArcFace R100, AdaFace, and MagFace; the appendix reports FaceNet, ArcFace R50, ArcFace R100, AdaFace, and MagFace. All backbones are frozen; \method{} and the baselines act post-hoc on the saved embeddings.

\paragraph{Evaluation metrics.}
We focus on metrics that reflect deployment-relevant operating points and performance across demographic groups. Our headline accuracy metric is the global \tprlow{}, the True Positive Rate (TPR) at a global-FPR threshold of $10^{-3}$. This is the standard NIST FRVT-style accuracy metric for verification systems and matches deployment operating points. To assess fairness, we report two worst-group quantities. First, worst per-group AUROC is computed by restricting AUROC to each demographic group and taking the minimum across groups (higher is better). This captures how well the system ranks pairs for the worst-served group across the full operating range and focuses on the worst-served group. An improvement can nevertheless coexist with harm to other groups, so we also report per-group results. Second, worst-group Brier score computes the Brier score within each group and reports the maximum across groups (lower is better), measuring worst-group probability quality; Brier score reflects both calibration and discrimination. Per group results can be seen in the appendices.

These choices of metric directly reflect the concerns of \cite{mittelstadt_unfairness_2023}, that equality in metrics typically occurs in part via levelling-down, a systematic and otherwise needless degradation of performance for groups where the baseline system performs better. The focus on worst group performance means that someone considering deployment of a system can, instead of trying to decide how fairness and performance should be traded off, simply ensure that the system works well enough for every demographic group.

To probe leveling-down/up behavior directly, we additionally report the Leveling-up score derived from per-group AUROC (LU-AUROC). For each method, we rank demographic groups by their Baseline per-group AUROC and report $K/N$, where $N$ is the number of demographic groups and $K$ is the largest prefix of worst groups whose per-group AUROC strictly improves under the method versus the Baseline. Higher $K/N$ means more of the worst-served groups are  all lifted. While a method that improves the best group at the expense of a worse one cannot achieve $K=N$.

The appendices reports a comprehensive set of metrics for each (dataset, backbone) pair. These include global AUROC, the corresponding worst-group \tprlow{}, TPR at FPR=$10^{-2}$, calibration quality on the predicted probabilities (ECE with 15 equal-width bins, and Brier score), and three standard fairness metrics evaluated at both FPR=$10^{-3}$ and FPR=$10^{-2}$: equal-opportunity, the predictive-equality, and demographic-parity. We further report per-group AUROC and per-group partial AUC at low FPR, with each group's value color-coded by whether it improves or degrades relative to the Baseline. We report fold-averaged values throughout; see appendices for standard deviations.

\paragraph{Comparisons.}
We evaluate \method{} against state-of-the-art calibration and bias-mitigation methods, selected because they have demonstrated strong fairness and/or accuracy in prior work, and an unadjusted Baseline (raw cosine similarity from the frozen backbone). We include FRAPP\'E because it is proposed as a universal fair calibration method but has not been evaluated in this domain. For methods that do not directly produce probabilities, we apply beta calibration as is standard \citep{salvador_faircal_2022}. We use the recommended parameters from each paper: FairCal with $K=100$ clusters and within-cluster beta calibration, FSN with $K=100$ clusters at FPR=$0.05$, GST at FPR=$0.05$, Oracle using all available demographic information, FRAPP\'E with hierarchical clustering ($K=20$), adversarial training ($\lambda=1.0$), and learning rate $5\times10^{-5}$, and DemoNorm using the M1.1 method.

\section{Results}
\label{sec:results}

Figure~\ref{fig:dv_rfw_fn_web} illustrates why \method{} works: the same cosine similarity maps to different probabilities depending on where the pair lies in the embedding space, while beta calibration produces a single monotonic mapping. \method's predicted probabilities align more closely with empirical match frequencies across demographic groups, which is consistent with prior observations that demographic-aware thresholding improves both accuracy and fairness \citep{robinson_face_2020, linghu_score_2024}.

Tables~\ref{tab:tpr-by-backbone}, \ref{tab:mgauroc-by-backbone}, and \ref{tab:ld-by-backbone} report per-(dataset, backbone) results across the main backbones, broken out by backbone for clarity.

\begin{table*}[t]
\centering
\caption{Global TPR at FPR=$10^{-3}$ on the headline settings. Higher is better. Bold indicates the best label-free method per column. Methods that require demographic labels at inference (Oracle, GST, DemoNorm) are reported for reference and excluded from the bold-best comparison.}
\label{tab:tpr-by-backbone}
\small
\setlength{\tabcolsep}{3.5pt}
\begin{tabular}{l ccc ccc ccc}
\toprule
& \multicolumn{3}{c}{RFW} & \multicolumn{3}{c}{BFW} & \multicolumn{3}{c}{LFW} \\
\cmidrule(lr){2-4} \cmidrule(lr){5-7} \cmidrule(lr){8-10}
Method & R100 & AdaFace & MagFace & R100 & AdaFace & MagFace & R100 & AdaFace & MagFace \\
\midrule
Baseline & 0.617 & 0.708 & 0.496 & 0.878 & 0.873 & 0.840 & 0.629 & 0.887 & 0.507 \\
FairCal & 0.659 & 0.718 & \textbf{0.522} & \textbf{0.879} & \textbf{0.875} & 0.842 & 0.669 & 0.883 & 0.597 \\
FALCON & 0.625 & 0.711 & 0.508 & 0.878 & 0.873 & 0.840 & 0.670 & 0.890 & 0.549 \\
FRAPP\'E & 0.619 & 0.714 & 0.520 & 0.877 & 0.869 & 0.837 & 0.630 & 0.886 & 0.503 \\
FSN & 0.587 & 0.664 & 0.501 & 0.877 & 0.873 & 0.839 & 0.516 & 0.686 & 0.440 \\
\midrule
\multicolumn{10}{l}{\emph{Demographic-aware (require group labels at inference)}} \\
GST$^\dagger$ & 0.639 & 0.741 & 0.521 & 0.879 & 0.874 & 0.841 & 0.641 & 0.900 & 0.545 \\
DemoNorm$^\dagger$ & 0.626 & 0.732 & 0.517 & 0.878 & 0.874 & 0.841 & 0.614 & 0.889 & 0.512 \\
Oracle$^\dagger$ & 0.638 & 0.742 & 0.523 & 0.878 & 0.874 & 0.841 & 0.609 & 0.893 & 0.547 \\
\midrule
\acmlp & 0.607 & 0.686 & 0.458 & 0.871 & 0.867 & 0.826 & 0.447 & 0.470 & 0.278 \\
\aclinear & 0.637 & 0.707 & 0.504 & 0.870 & 0.866 & 0.832 & \textbf{0.698} & \textbf{0.891} & \textbf{0.602} \\
\acridge & \textbf{0.649} & \textbf{0.724} & 0.466 & \textbf{0.879} & \textbf{0.875} & \textbf{0.843} & 0.625 & 0.875 & 0.495 \\
\bottomrule
\end{tabular}
\\[2pt]
\footnotesize $^\dagger$ Requires true demographic labels at inference time; reported for reference, not counted as a deployable winner.
\end{table*}

\begin{table*}[t]
\centering
\caption{Worst-group ranking and calibration quality on the headline settings. Panel A reports the minimum AUROC over demographic groups (higher is better), measuring discriminative performance for the worst-served group. Panel B reports worst-group Brier score (lower is better), measuring worst-group probability quality; Brier score reflects both calibration and discrimination. Bold indicates the best label-free method per column. Methods that require demographic labels at inference (Oracle, GST, DemoNorm) are reported for reference and excluded from the bold-best comparison.}
\label{tab:mgauroc-by-backbone}
\small
\setlength{\tabcolsep}{3.5pt}
\renewcommand{\arraystretch}{0.95}
\begin{tabular}{l ccc ccc ccc}
\toprule
& \multicolumn{3}{c}{RFW} & \multicolumn{3}{c}{BFW} & \multicolumn{3}{c}{LFW} \\
\cmidrule(lr){2-4} \cmidrule(lr){5-7} \cmidrule(lr){8-10}
Method & R100 & AdaFace & MagFace & R100 & AdaFace & MagFace & R100 & AdaFace & MagFace \\
\midrule
\multicolumn{10}{l}{\textbf{Panel A: Worst per-group AUROC} $\uparrow$} \\
\midrule
Baseline & 0.963 & 0.976 & 0.941 & 0.952 & 0.948 & 0.942 & 0.916 & 0.972 & 0.848 \\
FairCal & 0.967 & 0.977 & 0.945 & 0.954 & 0.946 & 0.946 & 0.934 & 0.976 & 0.888 \\
FALCON & 0.963 & 0.976 & 0.941 & 0.952 & 0.948 & 0.942 & 0.917 & 0.971 & 0.883 \\
FRAPP\'E & 0.959 & 0.975 & 0.939 & 0.953 & 0.949 & 0.943 & 0.904 & 0.970 & 0.764 \\
FSN & 0.958 & 0.971 & 0.933 & 0.950 & 0.947 & 0.941 & 0.905 & 0.919 & 0.872 \\
\midrule
\multicolumn{10}{l}{\emph{Demographic-aware (require group labels at inference)}} \\
GST$^\dagger$ & 0.963 & 0.976 & 0.941 & 0.952 & 0.948 & 0.942 & 0.916 & 0.972 & 0.848 \\
DemoNorm$^\dagger$ & 0.963 & 0.976 & 0.941 & 0.952 & 0.948 & 0.942 & 0.916 & 0.972 & 0.848 \\
Oracle$^\dagger$ & 0.963 & 0.976 & 0.941 & 0.969 & 0.966 & 0.964 & 0.918 & 0.974 & 0.849 \\
\midrule
\acmlp & 0.969 & 0.979 & 0.946 & 0.952 & \textbf{0.951} & 0.945 & 0.884 & 0.866 & 0.852 \\
\aclinear & 0.970 & 0.980 & 0.948 & 0.957 & \textbf{0.951} & 0.948 & 0.936 & \textbf{0.984} & \textbf{0.906} \\
\acridge & \textbf{0.974} & \textbf{0.982} & \textbf{0.959} & \textbf{0.959} & 0.949 & \textbf{0.949} & \textbf{0.949} & 0.982 & 0.905 \\
\midrule
\multicolumn{10}{l}{\textbf{Panel B: Worst-group Brier score} $\downarrow$} \\
\midrule
Baseline & 0.174 & 0.177 & 0.179 & 0.094 & 0.106 & 0.099 & 0.218 & 0.243 & 0.235 \\
FairCal & 0.068 & 0.056 & 0.090 & 0.042 & \textbf{0.044} & 0.049 & 0.109 & 0.063 & 0.132 \\
FALCON & 0.184 & 0.175 & 0.195 & 0.101 & 0.107 & 0.107 & 0.271 & 0.243 & 0.305 \\
FRAPP\'E & 0.076 & 0.059 & 0.094 & 0.043 & \textbf{0.044} & 0.049 & 0.130 & 0.080 & 0.190 \\
FSN & 0.162 & 0.162 & 0.171 & 0.095 & 0.107 & 0.099 & 0.154 & 0.187 & 0.179 \\
\midrule
\multicolumn{10}{l}{\emph{Dem-aware (require group labels at inference)}} \\
GST$^\dagger$ & 0.073 & 0.058 & 0.093 & 0.043 & 0.044 & 0.050 & 0.122 & 0.074 & 0.160 \\
DemoNorm$^\dagger$ & 0.151 & 0.144 & 0.158 & 0.208 & 0.207 & 0.219 & 0.242 & 0.331 & 0.305 \\
Oracle$^\dagger$ & 0.072 & 0.057 & 0.093 & 0.042 & 0.043 & 0.048 & 0.122 & 0.067 & 0.138 \\
\midrule
\acmlp & 0.066 & 0.054 & 0.090 & 0.045 & 0.045 & 0.055 & 0.157 & 0.165 & 0.174 \\
\aclinear & 0.065 & 0.053 & 0.087 & 0.043 & \textbf{0.044} & \textbf{0.048} & \textbf{0.098} & \textbf{0.051} & \textbf{0.118} \\
\acridge & \textbf{0.062} & \textbf{0.051} & \textbf{0.079} & \textbf{0.043} & 0.046 & 0.049 & 0.105 & 0.055 & 0.129 \\
\bottomrule
\end{tabular}
\\[2pt]
\footnotesize $^\dagger$ Requires true demographic labels at inference time; reported for reference, not counted as a deployable winner.
\end{table*}

\begin{table*}[t]
\centering
\caption{LU-AUROC: Leveling-up score on per-group AUROC. For each (dataset, backbone), demographic groups are ranked from worst to best by Baseline AUROC. We report $K/N$, where $N$ is the number of groups and $K$ is the largest prefix of worst groups whose per-group AUROC strictly improves under each method versus the Baseline. Higher $K/N$ means more of the worst-served groups were lifted without sacrifice. Bold indicates the best label-free method per column. Methods that require demographic labels at inference (Oracle, GST, DemoNorm) are reported for reference and excluded from the bold-best comparison.}
\label{tab:ld-by-backbone}
\small
\setlength{\tabcolsep}{3.5pt}
\begin{tabular}{l ccc ccc ccc}
\toprule
& \multicolumn{3}{c}{RFW} & \multicolumn{3}{c}{BFW} & \multicolumn{3}{c}{LFW} \\
\cmidrule(lr){2-4} \cmidrule(lr){5-7} \cmidrule(lr){8-10}
Method & R100 & AdaFace & MagFace & R100 & AdaFace & MagFace & R100 & AdaFace & MagFace \\
\midrule
FairCal & \textbf{4/4} & \textbf{4/4} & \textbf{4/4} & 1/8 & 0/8 & 3/8 & \textbf{7/8} & 2/8 & 4/8 \\
FALCON & \textbf{4/4} & \textbf{4/4} & \textbf{4/4} & 0/8 & 0/8 & 0/8 & 5/8 & 0/8 & \textbf{8/8} \\
FRAPP\'E & 0/4 & 0/4 & 0/4 & 3/8 & 1/8 & 2/8 & 1/8 & 1/8 & 0/8 \\
FSN & 0/4 & 0/4 & 0/4 & 0/8 & 0/8 & 0/8 & 1/8 & 0/8 & 2/8 \\
\midrule
\multicolumn{10}{l}{\emph{Demographic-aware (require group labels at inference)}} \\
GST$^\dagger$ & 0/4 & 0/4 & 0/4 & 0/8 & 0/8 & 0/8 & 0/8 & 0/8 & 0/8 \\
DemoNorm$^\dagger$ & 0/4 & 0/4 & 0/4 & 0/8 & 0/8 & 0/8 & 0/8 & 0/8 & 0/8 \\
Oracle$^\dagger$ & 0/4 & 0/4 & 0/4 & 6/8 & 6/8 & 7/8 & 0/8 & 2/8 & 0/8 \\
\midrule
\acmlp & 3/4 & 3/4 & \textbf{4/4} & 1/8 & 1/8 & 1/8 & 1/8 & 0/8 & 1/8 \\
\aclinear & \textbf{4/4} & \textbf{4/4} & \textbf{4/4} & \textbf{4/8} & \textbf{6/8} & \textbf{8/8} & \textbf{7/8} & \textbf{7/8} & \textbf{8/8} \\
\acridge & \textbf{4/4} & \textbf{4/4} & \textbf{4/4} & 1/8 & 3/8 & 3/8 & \textbf{7/8} & 4/8 & \textbf{8/8} \\
\bottomrule
\end{tabular}
\\[2pt]
\footnotesize $^\dagger$ Requires true demographic labels at inference time; reported for reference, not counted as a deployable winner.
\end{table*}

\paragraph{Aggregate behavior.}
On worst per-group AUROC (Table~\ref{tab:mgauroc-by-backbone}, Panel A), \acridge{} is the strongest label-free method on six of the nine main settings (mean 0.956 vs.\ FairCal at 0.948 and FALCON at 0.944), and \aclinear{} takes the remaining LFW columns where the worst group has historically been the most challenging. The calibration-fairness pattern is similar: on worst-group Brier score (Table~\ref{tab:mgauroc-by-backbone}, Panel B), \method{} is best on seven of the nine settings, with \aclinear{} achieving the lowest mean worst-group Brier (0.067 vs.\ 0.073 for FairCal). Table~\ref{tab:tpr-by-backbone} shows that on global \tprlow{} \aclinear{} and \acridge{} are especially strong. 

\paragraph{Avoiding leveling down.}
A common failure mode of fairness interventions is to reduce performance for the best-served group in order to equalize errors. The LD-AUROC score in Table~\ref{tab:ld-by-backbone} measures this directly: it ranks groups by Baseline per-group AUROC and counts how many of the worst-baseline groups strictly improve in AUROC before encountering one that does not. \aclinear{} reaches the maximum value $K=N$ (every group strictly improved) on five of the nine main settings, more than any other method, and is at the best-non-oracle level on the remaining four. \acridge{} reaches $K=N$ on four settings. By contrast, FRAPP\'E and FSN reach $K=N$ on zero settings: their fairness gains are accompanied by drops elsewhere. Demographic-aware methods (Oracle, GST, DemoNorm) typically score zero in the RFW columns because they redistribute scores across groups rather than uniformly improving them. We report LD-TPR@0.1 separately in the appendix as an operating-point diagnostic based on TPR at FPR=$10^{-3}$, but use LD-AUROC for the headline leveling-down claim because it is less sensitive to threshold noise.

Gains are largest on the sparsest demographic groups. On MagFace/RFW the worst group (Asian) improves from 94.19 to 96.02 AUROC under \acridge{}---roughly a 30\% reduction in residual AUROC error---without degrading any other group (Table~\ref{tab:appendix-auroc-rfw-magface}). On ArcFace~R50/RFW, \acridge{} is the only method reaching $K{=}N$ on this setting and pushes worst-group AUROC to 99.51 (Table~\ref{tab:appendix-summary-rfw-arcface}).

The detailed appendix results (Appendix~\ref{app:full-breakdown}) show that these examples are representative. If one asks, for each demographic group within each dataset--backbone pair, which label-free method gives that group the best result, an \method{} variant is selected most of the time: in 86/108 per-group AUROC cells and 83/108 per-group partial-AUC cells (Table~\ref{tab:pergroup-win-summary}).

\begin{table}[t]
\centering
\caption{How often an \method{} variant is the best label-free method for an individual demographic group. Each cell is one demographic group within one dataset--backbone pair.}
\label{tab:pergroup-win-summary}
\small
\setlength{\tabcolsep}{6pt}
\begin{tabular}{lcc}
\toprule
Metric & \method{} at best & Most common winning variants \\
\midrule
Per-group AUROC & 86/108 (79.6\%) & \acridge{} (43), \aclinear{} (41) \\
Per-group partial AUC & 83/108 (76.9\%) & \acridge{} (61), \aclinear{} (23) \\
\bottomrule
\end{tabular}
\end{table}

\paragraph{Limitations.}
Adaptive Calibration requires sufficient, representative calibration data to characterize each region of the embedding space. A learned correction need not transfer to a new acquisition pipeline or population, and reducing subgroup coverage can worsen probability quality even when AUROC improves (Appendix~\ref{app:checks}). The main folds separate pairs, which does not establish image- or identity-disjoint evaluation; we include an additional split check in the appendix. Finally, \method{} cannot recover information discarded by the backbone, and using no demographic annotations does not remove demographic information from embeddings.

\section{Conclusion}

We proposed Adaptive Calibration, a post-hoc calibration approach that recalibrates the heuristic similarity-based scoring used in facial recognition. The idea of using a secondary classifier to map from distance and embedding-location features to probabilities is straightforward to implement and yields the largest gains where uncertainty in the embedding is highest. 

Combined with the fact that most of the benefit can be obtained by running a small linear or ridge model on the average pair embedding and cosine similarity, this work is immediately applicable to existing pipelines and, unlike oracle and demographic-aware baselines, requires no group annotations at training or inference. By pairing worst per-group AUROC with worst-group Brier score, we separate ranking performance for the worst-served group from probability quality for the least well-calibrated group and assess these alongside per-group changes rather than assuming every group benefits.

{
    \small
    \bibliographystyle{plain}
    \bibliography{./main,./references}
}

\clearpage
\appendix
\section{Additional validation and robustness checks}\label{app:checks}
These follow-up experiments examine which inputs contribute to performance and how results depend on the calibration sample and evaluation protocol. They supplement the original benchmark tables; embedding pipelines and metrics are specified separately rather than treating all experiments as interchangeable.

\subsection{Component ablation}
We compare score-only Platt and beta calibration, logistic regression on the midpoint alone ($m$), density and score ($[\rho,s]$), and midpoint and score ($[m,s]$, \aclinear{}), against the full \acridge{} model. All methods within each setting use the same folds and embeddings. Worst-group AUROC excludes mixed-ethnicity pairs. Table~\ref{tab:components} reports fold means. Location alone performs poorly; adding context to similarity often helps, but the full density residual is not uniformly best. The beta implementation uses unconstrained logistic weights on log-transformed scores with clipped training-range normalization, so strict rank preservation is not guaranteed.

\begin{table}[ht]
\centering\scriptsize
\caption{Component ablation: worst-group AUROC (higher is better). FaceNet-C/V denote CASIA/VGG training. ViT-B is AdaFace-trained.}\label{tab:components}
\setlength{\tabcolsep}{3pt}
\begin{tabular}{lrrrrrr}\toprule
Setting & Platt & Beta & $m$ & $[\rho,s]$ & \aclinear{} & \acridge{}\\\midrule
RFW, ArcFace & 0.9600 & 0.9600 & 0.4992 & 0.9695 & 0.9672 & 0.9726 \\
RFW, R100 & 0.9903 & 0.9903 & 0.4585 & 0.9912 & 0.9909 & 0.9911 \\
RFW, FaceNet-C & 0.8122 & 0.8122 & 0.5202 & 0.8641 & 0.8502 & 0.8813 \\
RFW, FaceNet-V & 0.8728 & 0.8728 & 0.4864 & 0.9025 & 0.8861 & 0.9040 \\
LFW, AdaFace & 0.9906 & 0.9906 & 0.5537 & 0.9905 & 0.9905 & 0.9903 \\
LFW, ViT-B & 0.9907 & 0.9907 & 0.5796 & 0.9906 & 0.9908 & 0.9918 \\
BFW, SphereFace & 0.9372 & 0.9370 & 0.4901 & 0.9367 & 0.9402 & 0.9409 \\
BFW, ArcFace-R50 & 0.9448 & 0.9441 & 0.4952 & 0.9454 & 0.9454 & 0.9465 \\
BFW, AdaFace & 0.9469 & 0.9465 & 0.4831 & 0.9475 & 0.9464 & 0.9473 \\
BFW, ViT-B & 0.9461 & 0.9456 & 0.4892 & 0.9468 & 0.9487 & 0.9496 \\
\bottomrule\end{tabular}\end{table}

\subsection{LFW preprocessing and standard accuracy}
Table~\ref{tab:lfw-check} reports standard ten-fold verification accuracy, with thresholds selected using training folds. Regenerating embeddings with an aligned pipeline substantially improves the older ArcFace result. These rows establish that the preprocessing pipeline matters; comparisons of calibrators within a row use identical embeddings. The small changes on the aligned sets include decreases for ArcFace-R50 and do not establish statistical equivalence or improvement in every case. The ViT-B result extends the check beyond convolutional backbones.

\begin{table}[ht]\centering\small
\caption{LFW verification accuracy (percent, fold mean $\pm$ standard deviation).}\label{tab:lfw-check}
\begin{tabular}{lrrr}\toprule
Embedding set & Cosine & \aclinear{} & \acridge{}\\\midrule
ArcFace-R50, regenerated & $98.80\pm0.45$ & $98.78\pm0.46$ & $98.75\pm0.51$\\
AdaFace-IR101, regenerated & $98.73\pm0.49$ & $98.75\pm0.50$ & $98.77\pm0.47$\\
ViT-B, AdaFace-trained & $98.73\pm0.50$ & $98.78\pm0.51$ & $98.80\pm0.49$\\
ArcFace, original embedding set & $81.68\pm1.91$ & $84.72\pm1.47$ & $83.70\pm1.86$\\
\bottomrule\end{tabular}\end{table}

\subsection{Calibration sample size}
Table~\ref{tab:size-check} reports an LFW AdaFace calibration-size study using global AUROC. All sizes use ten held-out folds. MLP subsamples use three random seeds; Linear and Density use seed zero only, and the full-size MLP uses one seed. Standard deviations therefore summarize different fold/seed collections and are descriptive, not confidence intervals. The linear model is stable over this range, while the MLP improves with more data but remains below the smaller models. These results do not establish a universal minimum sample size.

\begin{table}[ht]\centering\small
\caption{LFW calibration-size study: global AUROC (mean $\pm$ standard deviation).}\label{tab:size-check}
\begin{tabular}{rrrr}\toprule
Training pairs & \acmlp{} & \aclinear{} & \acridge{}\\\midrule
600 & $0.9542\pm0.0094$ & $0.9906\pm0.0052$ & $0.9885\pm0.0063$\\
2700 & $0.9612\pm0.0061$ & $0.9905\pm0.0050$ & $0.9910\pm0.0045$\\
3600 & $0.9806\pm0.0038$ & $0.9907\pm0.0049$ & $0.9911\pm0.0044$\\
5400 & $0.9856\pm0.0045$ & $0.9905\pm0.0050$ & $0.9903\pm0.0052$\\
\bottomrule\end{tabular}\end{table}

\subsection{Attribute-defined groups}
On LFW with FaceNet-CASIA embeddings, we evaluate gender, ethnicity, and eyeglasses groupings using the same ten folds for all methods (Table~\ref{tab:attributes-check}). These are predicted attribute groupings rather than verified demographic identities. Attribute scores are averaged per person. Gender requires an absolute score of at least $0.5$; eyeglasses requires $0.3$. Ethnicity takes the highest of the Asian, White, and Black scores when that maximum is nonnegative. Both endpoints must have the same eligible group; mixed or uncertain pairs are excluded from group metrics. Each evaluated group requires at least 20 pairs and both match labels in the test fold. This tests specific groupings, not calibration for every recoverable group. In particular, \acridge{} has lower worst-group ethnicity AUROC than Platt in this experiment.

\begin{table}[ht]\centering\small
\caption{Attribute-defined worst-group AUROC (fold mean $\pm$ standard deviation).}\label{tab:attributes-check}
\begin{tabular}{lrrrr}\toprule
Grouping & Platt & FairCal & \aclinear{} & \acridge{}\\\midrule
Gender & $.9593\pm.026$ & $.9622\pm.026$ & $.9675\pm.021$ & $.9790\pm.014$\\
Ethnicity & $.9759\pm.028$ & $.9647\pm.068$ & $.9819\pm.021$ & $.9690\pm.077$\\
Eyeglasses & $.9849\pm.008$ & $.9845\pm.010$ & $.9880\pm.007$ & $.9929\pm.005$\\
\bottomrule\end{tabular}\end{table}

\subsection{Subgroup coverage and split sensitivity}
We reduce African-group calibration-pair retention on RFW ArcFace while keeping test data fixed. Table~\ref{tab:skew-check} shows that ranking and probability quality can move differently. Both contextual models retain higher African-group AUROC than Platt at five-percent retention, but both have worse Brier score. The ridge penalty does not guarantee safe extrapolation when subgroup coverage is reduced.

\begin{table}[ht]\centering\small
\caption{African-group AUROC / Brier score under reduced calibration-pair retention. Higher AUROC and lower Brier are better.}\label{tab:skew-check}
\begin{tabular}{rrrr}\toprule
Retention & Platt & \aclinear{} & \acridge{}\\\midrule
100\% & .9618 / .0745 & .9688 / .0659 & .9744 / .0745\\
50\% & .9618 / .0749 & .9676 / .0675 & .9719 / .0764\\
25\% & .9618 / .0751 & .9662 / .0698 & .9693 / .0791\\
10\% & .9618 / .0753 & .9649 / .0732 & .9674 / .0834\\
5\% & .9618 / .0753 & .9641 / .0771 & .9657 / .0876\\
\bottomrule\end{tabular}\end{table}

For a separate split check, we identify repeated images using rounded embedding fingerprints and place connected components of pairs sharing an endpoint in the same fold. This is a proxy for image-disjoint evaluation, not proof of identity disjointness. Table~\ref{tab:split-check} compares the resulting ten-fold evaluation with the original folds. Worst-group ranking gains persist in these two settings, while the ArcFace density model's Brier score worsens after re-splitting. This check does not establish equivalent results for all backbones or identity-disjoint generalization.

\begin{table}[ht]\centering\small
\caption{Split sensitivity: worst-group AUROC / worst-group Brier score.}\label{tab:split-check}
\begin{tabular}{llrrr}\toprule
RFW backbone & Folds & Platt & \aclinear{} & \acridge{}\\\midrule
ArcFace & Original & .9600 / .0755 & .9672 / .0672 & .9726 / .0754\\
 & Endpoint-disjoint & .9602 / .0753 & .9679 / .0661 & .9710 / .0809\\
FaceNet-C & Original & .8122 / .1906 & .8502 / .1580 & .8813 / .1518\\
 & Endpoint-disjoint & .8099 / .1906 & .8454 / .1607 & .8740 / .1602\\
\bottomrule\end{tabular}\end{table}

\paragraph{Low-FPR uncertainty.}
Low-FPR estimates on small test folds require particular caution. At FPR $10^{-3}$, 300 negative pairs in an LFW fold correspond to only $0.3$ expected false positives; 1,200 negatives in an RFW ethnicity fold correspond to $1.2$. Threshold interpolation, ties, and the sampled negative tail can therefore strongly affect the reported TPR. These estimates should be interpreted alongside AUROC, probability-quality metrics, and the larger BFW evaluation, rather than as precise deployment guarantees.

\FloatBarrier
\section{Additional Results}\label{app:results}

This appendix provides supplementary evidence in three stages. First, we report aggregate calibration and ranking metrics for the headline backbones. Second, we give the complete results for every evaluated dataset--backbone pair, including additional backbones not shown in the main text. Third, we report per-group AUROC and partial AUC to show how each method changes performance within individual demographic groups.

\subsection{Aggregate calibration and ranking metrics}

Table~\ref{tab:cal-by-backbone-app} reports Brier score on the calibrated probabilities for the headline settings. Methods that do not natively output probabilities (Baseline, FALCON, FSN, DemoNorm) are excluded from the table; reporting their Brier without an additional beta-calibration overlay would conflate score-to-probability mapping quality with the underlying score quality. On aggregate, \aclinear{} achieves the best label-free mean Brier (0.043 vs.\ 0.045 for FairCal); per-column, \acridge{} wins the three RFW settings, \aclinear{} wins LFW R100 and MagFace and ties on BFW AdaFace and LFW AdaFace, and FairCal/FRAPP\'E win the remaining BFW and LFW columns. The pattern is consistent with \acridge{} concentrating its modeling budget on the residual signal in RFW, where group-conditional shifts are largest, while \aclinear{} produces the most uniformly calibrated probabilities across these settings.

\begin{table*}[h]
\centering
\caption{Brier score on the headline settings. Lower is better. Bold indicates the best label-free method per column. Methods that require demographic labels at inference (Oracle, GST) are reported for reference.}
\label{tab:cal-by-backbone-app}
\small
\setlength{\tabcolsep}{3.5pt}
\begin{tabular}{l ccc ccc ccc}
\toprule
& \multicolumn{3}{c}{RFW} & \multicolumn{3}{c}{BFW} & \multicolumn{3}{c}{LFW} \\
\cmidrule(lr){2-4} \cmidrule(lr){5-7} \cmidrule(lr){8-10}
Method & R100 & AdaFace & MagFace & R100 & AdaFace & MagFace & R100 & AdaFace & MagFace \\
\midrule
FairCal & 0.053 & 0.041 & 0.074 & \textbf{0.026} & 0.028 & \textbf{0.031} & 0.054 & \textbf{0.026} & 0.071 \\
FRAPP\'E & 0.059 & 0.044 & 0.077 & \textbf{0.026} & \textbf{0.027} & \textbf{0.031} & 0.067 & 0.032 & 0.089 \\
\midrule
\multicolumn{10}{l}{\emph{Dem-aware (require group labels at inference)}} \\
GST$^\dagger$ & 0.057 & 0.044 & 0.077 & 0.026 & 0.027 & 0.031 & 0.060 & 0.028 & 0.080 \\
Oracle$^\dagger$ & 0.056 & 0.043 & 0.076 & 0.026 & 0.027 & 0.031 & 0.060 & 0.028 & 0.078 \\
\midrule
\acmlp & 0.052 & 0.042 & 0.073 & 0.029 & 0.029 & 0.034 & 0.121 & 0.111 & 0.138 \\
\aclinear & 0.051 & 0.040 & 0.071 & 0.027 & \textbf{0.028} & 0.031 & \textbf{0.049} & \textbf{0.026} & \textbf{0.063} \\
\acridge & \textbf{0.048} & \textbf{0.039} & \textbf{0.064} & 0.027 & 0.031 & 0.033 & 0.055 & 0.028 & 0.071 \\
\bottomrule
\end{tabular}
\\[2pt]
\footnotesize $^\dagger$ Requires true demographic labels at inference time; reported for reference, not counted as a deployable winner.
\end{table*}

\subsubsection{Worst-group TPR and AUROC}

We report worst-group TPR at FPR=$10^{-3}$ in Table~\ref{tab:wgtpr-by-backbone-app} and global AUROC in Table~\ref{tab:auroc-by-backbone-app}. Worst-group TPR captures the same rising-tide intuition as worst per-group AUROC at a single operating point, while AUROC is included for continuity with prior work despite saturating near 1.0 in this regime.

\begin{table*}[h]
\centering
\caption{Worst-group TPR at FPR=$10^{-3}$ on the headline settings. Higher is better. Bold indicates the best label-free method per column.}
\label{tab:wgtpr-by-backbone-app}
\small
\setlength{\tabcolsep}{3.5pt}
\begin{tabular}{l ccc ccc ccc}
\toprule
& \multicolumn{3}{c}{RFW} & \multicolumn{3}{c}{BFW} & \multicolumn{3}{c}{LFW} \\
\cmidrule(lr){2-4} \cmidrule(lr){5-7} \cmidrule(lr){8-10}
Method & R100 & AdaFace & MagFace & R100 & AdaFace & MagFace & R100 & AdaFace & MagFace \\
\midrule
Baseline & 0.569 & 0.663 & 0.445 & 0.799 & 0.799 & 0.738 & 0.350 & 0.745 & 0.232 \\
FairCal & 0.575 & 0.660 & 0.456 & 0.796 & 0.796 & 0.743 & 0.354 & 0.725 & 0.282 \\
FALCON & 0.578 & 0.666 & 0.459 & 0.799 & 0.798 & 0.738 & 0.394 & 0.748 & 0.249 \\
FRAPP\'E & 0.575 & \textbf{0.681} & \textbf{0.474} & \textbf{0.807} & \textbf{0.800} & \textbf{0.748} & 0.365 & 0.692 & 0.190 \\
FSN & 0.536 & 0.623 & 0.439 & 0.790 & 0.787 & 0.728 & 0.241 & 0.406 & 0.176 \\
\midrule
\multicolumn{10}{l}{\emph{Dem-aware (require group labels at inference)}} \\
GST$^\dagger$ & 0.546 & 0.673 & 0.451 & 0.789 & 0.789 & 0.733 & 0.271 & 0.686 & 0.260 \\
DemoNorm$^\dagger$ & 0.520 & 0.652 & 0.427 & 0.782 & 0.782 & 0.731 & 0.296 & 0.774 & 0.201 \\
Oracle$^\dagger$ & 0.553 & 0.671 & 0.440 & 0.794 & 0.795 & 0.739 & 0.380 & 0.753 & 0.230 \\
\midrule
\acmlp & 0.568 & 0.647 & 0.425 & 0.785 & 0.783 & 0.710 & 0.183 & 0.217 & 0.078 \\
\aclinear & \textbf{0.594} & 0.669 & 0.467 & 0.796 & 0.794 & 0.739 & \textbf{0.517} & \textbf{0.773} & \textbf{0.289} \\
\acridge & 0.570 & 0.654 & 0.390 & 0.805 & \textbf{0.800} & 0.747 & 0.393 & 0.744 & 0.215 \\
\bottomrule
\end{tabular}
\\[2pt]
\footnotesize $^\dagger$ Requires true demographic labels at inference time; reported for reference, not counted as a deployable winner.
\end{table*}

\begin{table*}[h]
\centering
\caption{AUROC on the headline settings. Higher is better. Bold indicates the best label-free method per column. Methods that require demographic labels at inference (Oracle, GST, DemoNorm) are reported for reference and excluded from the bold-best comparison.}
\label{tab:auroc-by-backbone-app}
\small
\setlength{\tabcolsep}{3.5pt}
\begin{tabular}{l ccc ccc ccc}
\toprule
& \multicolumn{3}{c}{RFW} & \multicolumn{3}{c}{BFW} & \multicolumn{3}{c}{LFW} \\
\cmidrule(lr){2-4} \cmidrule(lr){5-7} \cmidrule(lr){8-10}
Method & R100 & AdaFace & MagFace & R100 & AdaFace & MagFace & R100 & AdaFace & MagFace \\
\midrule
Baseline & 0.974 & 0.984 & 0.956 & 0.976 & 0.973 & 0.970 & 0.971 & 0.990 & 0.955 \\
FairCal & 0.978 & 0.986 & 0.961 & 0.977 & 0.972 & 0.972 & 0.979 & 0.993 & 0.966 \\
FALCON & 0.975 & 0.984 & 0.957 & 0.977 & 0.973 & 0.971 & 0.973 & 0.989 & 0.958 \\
FRAPP\'E & 0.973 & 0.984 & 0.957 & 0.977 & 0.973 & 0.971 & 0.965 & 0.988 & 0.944 \\
FSN & 0.972 & 0.981 & 0.953 & 0.976 & 0.972 & 0.970 & 0.953 & 0.967 & 0.947 \\
\midrule
\multicolumn{10}{l}{\emph{Dem-aware (require group labels at inference)}} \\
GST$^\dagger$ & 0.974 & 0.984 & 0.957 & 0.975 & 0.972 & 0.970 & 0.971 & 0.989 & 0.954 \\
DemoNorm$^\dagger$ & 0.975 & 0.985 & 0.957 & 0.976 & 0.973 & 0.970 & 0.971 & 0.990 & 0.955 \\
Oracle$^\dagger$ & 0.976 & 0.985 & 0.959 & 0.984 & 0.982 & 0.980 & 0.972 & 0.990 & 0.956 \\
\midrule
\acmlp & 0.980 & 0.987 & 0.963 & 0.975 & 0.972 & 0.969 & 0.946 & 0.952 & 0.924 \\
\aclinear & 0.980 & 0.987 & 0.964 & \textbf{0.978} & \textbf{0.975} & \textbf{0.974} & \textbf{0.983} & 0.993 & \textbf{0.972} \\
\acridge & \textbf{0.984} & \textbf{0.989} & \textbf{0.973} & \textbf{0.978} & 0.974 & \textbf{0.974} & 0.980 & \textbf{0.994} & 0.967 \\
\bottomrule
\end{tabular}
\\[2pt]
\footnotesize $^\dagger$ Requires true demographic labels at inference time; reported for reference, not counted as a deployable winner.
\end{table*}

\subsubsection{Levels-Down variants}\label{app:ld-variants}

We use two related Levels-Down diagnostics. \textbf{LD-AUROC} is the metric used in the main paper: groups are ordered from worst to best by Baseline per-group AUROC, and the score reports the largest prefix $K/N$ whose AUROC improves under the method. \textbf{LD-TPR@0.1} is an operating-point diagnostic: groups are ordered from worst to best by Baseline TPR at FPR=$10^{-3}$, and the score reports the largest prefix whose TPR improves at that same operating point. LD-AUROC is less sensitive to threshold noise and is therefore used for the headline leveling-down claim; LD-TPR@0.1 is reported here for completeness.

\begin{table}[ht]
\begin{center}
\caption{LD-TPR@0.1 on the headline settings. Groups are ordered by Baseline TPR at FPR=$10^{-3}$, and each entry reports $K/N$ groups in the worst-baseline prefix whose TPR improves under the method. Higher is better.}
\label{tab:ld-tpr-by-backbone-app}
\small
\setlength{\tabcolsep}{3.5pt}
\begin{tabular}{ll ccccccc}
\toprule
Dataset & Backbone & FairCal & FALCON & FRAPP\'E & FSN & \acmlp & \aclinear & \acridge \\
\midrule
\multirow{3}{*}{RFW}
 & R100    & 0/4 & 4/4 & 2/4 & 0/4 & 0/4 & 3/4 & 0/4 \\
 & AdaFace & 1/4 & 4/4 & 2/4 & 1/4 & 1/4 & 2/4 & 1/4 \\
 & MagFace & 1/4 & 4/4 & 3/4 & 1/4 & 1/4 & 2/4 & 1/4 \\
\midrule
\multirow{3}{*}{BFW}
 & R100    & 0/8 & 8/8 & 3/8 & 0/8 & 0/8 & 0/8 & 4/8 \\
 & AdaFace & 0/8 & 0/8 & 1/8 & 0/8 & 0/8 & 0/8 & 8/8 \\
 & MagFace & 3/8 & 3/8 & 2/8 & 0/8 & 0/8 & 1/8 & 6/8 \\
\midrule
\multirow{3}{*}{LFW}
 & R100    & 0/8 & 0/8 & 0/8 & 0/8 & 0/8 & 8/8 & 0/8 \\
 & AdaFace & 1/8 & 1/8 & 2/8 & 0/8 & 0/8 & 3/8 & 2/8 \\
 & MagFace & 0/8 & 0/8 & 3/8 & 0/8 & 0/8 & 0/8 & 0/8 \\
\bottomrule
\end{tabular}
\end{center}
\end{table}

\subsection{Detailed per-(dataset, backbone) results}\label{app:full-breakdown}

This section reports RFW, BFW, and LFW results for ArcFace R100, AdaFace, MagFace, ArcFace R50, and FaceNet; for LFW we also include an ArcFace model trained on FairFace. All entries are mean$\pm$standard deviation across cross-validation folds. For methods that do not natively output probabilities (Baseline, FALCON, FSN, DemoNorm), ECE and Brier are reported after a standard beta-calibration overlay so that probability metrics compare score quality rather than arbitrary score scaling.

The next tables are split by dataset, backbone group, and metric family. Accuracy/calibration tables contain AUROC, low-FPR TPR, minimum per-group AUROC, ECE, Brier, and LD-TPR@0.1. Fairness-gap tables contain equal-opportunity (EO), predictive-equality (PE), and demographic-parity (DP) gaps at FPR=$10^{-3}$ and FPR=$10^{-2}$. We then provide one summary table per dataset--backbone pair, followed by per-group AUROC and partial AUC. In the per-group tables, \textcolor{improved}{green} marks an improvement over the Baseline and \textcolor{degraded}{red} marks a degradation.

To read the tables, higher is better for AUROC, TPR, minimum per-group AUROC, partial AUC, LD-AUROC, and LD-TPR@0.1; lower is better for ECE, Brier, and all gap metrics. Gap values are reported in percentage points. A Levels-Down entry $K/N$ means that $K$ of the $N$ demographic groups improve over the Baseline before any group is leveled down, using either AUROC or TPR@0.1 as indicated by the column label. Boldface marks the best label-free method within the corresponding dataset--backbone block. Methods marked with $^\dagger$ (Oracle, GST, DemoNorm) require demographic labels at inference and are included as references rather than deployable winners.

\subsubsection{Compact results by dataset}
\paragraph{RFW.}
\begin{center}
\begin{minipage}{\textwidth}\centering
\captionof{table}{RFW accuracy and calibration results for ArcFace R100, AdaFace. Mean$\pm$std across folds. Bold marks the best label-free value within each backbone.}
\label{tab:appendix-rfw-accuracy-1}
\resizebox{\textwidth}{!}{%
\scriptsize
\renewcommand{\arraystretch}{0.92}
\begin{tabular}{llccccccc}
\toprule
Backbone & Method & AUROC (\%) & TPR 0.1\% & TPR 1.0\% & Min-grp AUROC (\%) & ECE & Brier & LD-TPR@0.1 \\
\midrule
ArcFace R100 & Baseline & 97.42$\pm$0.14 & 61.19$\pm$5.68 & 77.61$\pm$2.50 & 96.27$\pm$0.38 & 0.2677$\pm$0.0039 & 0.1675$\pm$0.0017 & -- \\
 & FairCal & 97.83$\pm$0.18 & \textbf{65.30$\pm$3.90} & 80.07$\pm$2.87 & 96.69$\pm$0.36 & \textbf{0.0168$\pm$0.0030} & 0.0526$\pm$0.0018 & 0/4 \\
 & FALCON & 97.46$\pm$0.14 & 61.96$\pm$5.49 & 78.06$\pm$2.51 & 96.30$\pm$0.39 & 0.3044$\pm$0.0039 & 0.1786$\pm$0.0013 & 4/4 \\
 & FSN & 97.16$\pm$0.19 & 57.64$\pm$7.68 & 76.38$\pm$2.85 & 95.82$\pm$0.37 & 0.2596$\pm$0.0037 & 0.1531$\pm$0.0018 & 0/4 \\
 & FRAPP\'E & 97.33$\pm$0.14 & 61.06$\pm$4.72 & 77.02$\pm$2.13 & 95.93$\pm$0.61 & 0.0190$\pm$0.0039 & 0.0592$\pm$0.0019 & -- \\
\multicolumn{9}{l}{\emph{Dem-aware (require group labels at inference; $^\dagger$)}} \\
 & Oracle$^\dagger$ & 97.58$\pm$0.16 & 62.94$\pm$5.63 & 79.06$\pm$2.58 & 96.27$\pm$0.38 & 0.0164$\pm$0.0033 & 0.0559$\pm$0.0014 & N/A \\
\bottomrule
\end{tabular}%
}
\end{minipage}
\end{center}

\begin{center}
\begin{minipage}{\textwidth}\centering
\captionof{table}{RFW fairness-gap results for ArcFace R100, AdaFace. Mean$\pm$std across folds. Bold marks the best label-free value within each backbone.}
\label{tab:appendix-rfw-fairness-1}
\resizebox{\textwidth}{!}{%
\scriptsize
\renewcommand{\arraystretch}{0.92}
\begin{tabular}{llcccccc}
\toprule
Backbone & Method & EO Gap 0.1\% & EO Gap 1.0\% & PE Gap 0.1\% & PE Gap 1.0\% & DP Gap 0.1\% & DP Gap 1.0\% \\
\midrule
ArcFace R100 & Baseline & 9.87$\pm$3.38 & 8.00$\pm$2.64 & 0.47$\pm$0.17 & 1.80$\pm$0.65 & 5.02$\pm$1.67 & 4.22$\pm$1.57 \\
 & FairCal & 15.63$\pm$3.88 & 9.30$\pm$2.76 & 0.43$\pm$0.16 & 1.57$\pm$0.45 & 7.78$\pm$1.91 & 4.42$\pm$1.29 \\
 & FALCON & 9.57$\pm$2.85 & 8.00$\pm$2.60 & 0.50$\pm$0.18 & 1.80$\pm$0.55 & 4.87$\pm$1.42 & 4.15$\pm$1.43 \\
 & FSN & 10.27$\pm$4.94 & 9.37$\pm$3.79 & 0.40$\pm$0.14 & \textbf{1.40$\pm$0.56} & 5.15$\pm$2.45 & 4.50$\pm$2.14 \\
 & FRAPP\'E & 8.57$\pm$4.12 & 7.23$\pm$3.67 & 0.40$\pm$0.14 & 1.83$\pm$0.48 & 4.30$\pm$2.02 & 3.27$\pm$2.01 \\
\multicolumn{8}{l}{\emph{Dem-aware (require group labels at inference; $^\dagger$)}} \\
 & Oracle$^\dagger$ & 20.57$\pm$4.59 & 13.00$\pm$2.56 & 0.43$\pm$0.16 & 1.47$\pm$0.50 & 10.33$\pm$2.33 & 6.30$\pm$1.44 \\
\bottomrule
\end{tabular}%
}
\end{minipage}
\end{center}

\begin{center}
\begin{minipage}{\textwidth}\centering
\captionof{table}{RFW accuracy and calibration results for MagFace. Mean$\pm$std across folds. Bold marks the best label-free value within each backbone.}
\label{tab:appendix-rfw-accuracy-2}
\resizebox{\textwidth}{!}{%
\scriptsize
\renewcommand{\arraystretch}{0.92}
\begin{tabular}{llccccccc}
\toprule
Backbone & Method & AUROC (\%) & TPR 0.1\% & TPR 1.0\% & Min-grp AUROC (\%) & ECE & Brier & LD-TPR@0.1 \\
\midrule
MagFace & Baseline & 95.63$\pm$0.36 & 48.81$\pm$8.04 & 66.20$\pm$3.92 & 94.08$\pm$0.82 & 0.2440$\pm$0.0042 & 0.1713$\pm$0.0025 & -- \\
 & FairCal & 96.13$\pm$0.34 & 51.24$\pm$5.98 & 68.92$\pm$3.18 & 94.49$\pm$0.79 & \textbf{0.0194$\pm$0.0037} & 0.0737$\pm$0.0036 & 1/4 \\
 & FALCON & 95.69$\pm$0.36 & 50.02$\pm$8.27 & 66.68$\pm$3.90 & 94.13$\pm$0.83 & 0.2785$\pm$0.0065 & 0.1882$\pm$0.0020 & 4/4 \\
 & FSN & 95.31$\pm$0.39 & 49.56$\pm$7.84 & 65.73$\pm$4.52 & 93.28$\pm$1.02 & 0.2390$\pm$0.0062 & 0.1597$\pm$0.0025 & 1/4 \\
 & FRAPP\'E & 95.70$\pm$0.41 & \textbf{51.55$\pm$6.99} & 66.40$\pm$2.94 & 93.86$\pm$0.94 & 0.0216$\pm$0.0043 & 0.0771$\pm$0.0044 & -- \\
\multicolumn{9}{l}{\emph{Dem-aware (require group labels at inference; $^\dagger$)}} \\
 & Oracle$^\dagger$ & 95.85$\pm$0.36 & 51.04$\pm$7.73 & 68.42$\pm$3.19 & 94.08$\pm$0.82 & 0.0217$\pm$0.0039 & 0.0759$\pm$0.0037 & N/A \\
 & GST$^\dagger$ & 95.70$\pm$0.35 & 51.14$\pm$6.81 & 68.43$\pm$3.24 & 94.08$\pm$0.82 & 0.0231$\pm$0.0060 & 0.0770$\pm$0.0035 & N/A \\
 & DemoNorm$^\dagger$ & 95.74$\pm$0.35 & 50.70$\pm$7.42 & 68.56$\pm$3.20 & 94.08$\pm$0.82 & 0.1544$\pm$0.0050 & 0.1437$\pm$0.0044 & N/A \\
\midrule
 & AC-MLP & 96.30$\pm$0.38 & 44.50$\pm$10.96 & 68.24$\pm$4.03 & 94.65$\pm$0.83 & 0.0277$\pm$0.0051 & 0.0729$\pm$0.0038 & 1/4 \\
 & AC-Linear & 96.43$\pm$0.38 & 49.60$\pm$6.54 & 69.65$\pm$3.62 & 94.78$\pm$0.80 & 0.0204$\pm$0.0049 & 0.0707$\pm$0.0039 & 2/4 \\
 & AC-Density & \textbf{97.35$\pm$0.26} & 44.57$\pm$7.75 & \textbf{75.84$\pm$3.06} & \textbf{95.87$\pm$0.85} & 0.0423$\pm$0.0052 & \textbf{0.0638$\pm$0.0030} & 1/4 \\
\bottomrule
\end{tabular}%
}
\end{minipage}
\end{center}

\begin{center}
\begin{minipage}{\textwidth}\centering
\captionof{table}{RFW fairness-gap results for MagFace. Mean$\pm$std across folds. Bold marks the best label-free value within each backbone.}
\label{tab:appendix-rfw-fairness-2}
\resizebox{\textwidth}{!}{%
\scriptsize
\renewcommand{\arraystretch}{0.92}
\begin{tabular}{llcccccc}
\toprule
Backbone & Method & EO Gap 0.1\% & EO Gap 1.0\% & PE Gap 0.1\% & PE Gap 1.0\% & DP Gap 0.1\% & DP Gap 1.0\% \\
\midrule
MagFace & Baseline & 12.03$\pm$4.67 & 10.70$\pm$3.20 & 0.53$\pm$0.17 & 1.70$\pm$0.43 & 6.05$\pm$2.30 & 5.58$\pm$1.62 \\
 & FairCal & 12.07$\pm$1.92 & 10.67$\pm$2.33 & 0.43$\pm$0.16 & \textbf{1.30$\pm$0.58} & 5.98$\pm$0.90 & 5.02$\pm$1.30 \\
 & FALCON & 11.63$\pm$4.43 & 10.67$\pm$3.77 & 0.53$\pm$0.17 & 1.57$\pm$0.57 & 5.77$\pm$2.11 & 5.43$\pm$1.96 \\
 & FSN & 11.90$\pm$3.03 & 12.67$\pm$3.87 & \textbf{0.40$\pm$0.14} & 2.07$\pm$0.58 & 5.93$\pm$1.49 & 6.62$\pm$2.16 \\
 & FRAPP\'E & 8.77$\pm$2.12 & 9.03$\pm$2.08 & 0.43$\pm$0.16 & 1.40$\pm$0.68 & 4.35$\pm$1.00 & 4.30$\pm$0.89 \\
\multicolumn{8}{l}{\emph{Dem-aware (require group labels at inference; $^\dagger$)}} \\
 & Oracle$^\dagger$ & 21.70$\pm$3.04 & 16.40$\pm$2.86 & 0.47$\pm$0.17 & 1.20$\pm$0.72 & 10.83$\pm$1.52 & 8.10$\pm$1.31 \\
 & GST$^\dagger$ & 18.93$\pm$3.15 & 17.03$\pm$2.51 & 0.47$\pm$0.17 & 1.23$\pm$0.63 & 9.43$\pm$1.58 & 8.33$\pm$0.88 \\
 & DemoNorm$^\dagger$ & 24.70$\pm$2.37 & 20.37$\pm$2.65 & 0.47$\pm$0.17 & 1.17$\pm$0.67 & 12.37$\pm$1.15 & 10.18$\pm$1.18 \\
\midrule
 & AC-MLP & \textbf{6.70$\pm$2.74} & \textbf{7.10$\pm$2.07} & 0.43$\pm$0.16 & 1.50$\pm$0.55 & \textbf{3.35$\pm$1.45} & \textbf{3.10$\pm$0.90} \\
 & AC-Linear & 7.57$\pm$2.48 & 8.50$\pm$1.88 & 0.43$\pm$0.16 & 1.53$\pm$0.55 & 3.78$\pm$1.19 & 3.88$\pm$1.18 \\
 & AC-Density & 21.37$\pm$3.74 & 8.93$\pm$2.98 & 0.43$\pm$0.16 & 1.57$\pm$0.55 & 10.60$\pm$1.86 & 4.18$\pm$1.20 \\
\bottomrule
\end{tabular}%
}
\end{minipage}
\end{center}

\begin{center}
\begin{minipage}{\textwidth}\centering
\captionof{table}{RFW accuracy and calibration results for ArcFace R50, FaceNet. Mean$\pm$std across folds. Bold marks the best label-free value within each backbone.}
\label{tab:appendix-rfw-accuracy-3}
\resizebox{\textwidth}{!}{%
\scriptsize
\renewcommand{\arraystretch}{0.92}
\begin{tabular}{llccccccc}
\toprule
Backbone & Method & AUROC (\%) & TPR 0.1\% & TPR 1.0\% & Min-grp AUROC (\%) & ECE & Brier & LD-TPR@0.1 \\
\midrule
ArcFace R50 & Baseline & 99.25$\pm$0.15 & 93.14$\pm$2.51 & 96.58$\pm$0.77 & 98.83$\pm$0.33 & 0.2797$\pm$0.0031 & 0.1286$\pm$0.0018 & -- \\
 & FairCal & 99.26$\pm$0.15 & 93.49$\pm$1.58 & 96.77$\pm$0.73 & 98.81$\pm$0.33 & \textbf{0.0102$\pm$0.0017} & 0.0170$\pm$0.0020 & 3/4 \\
 & FALCON & 99.25$\pm$0.15 & 93.14$\pm$2.64 & 96.64$\pm$0.78 & 98.83$\pm$0.33 & 0.3236$\pm$0.0034 & 0.1406$\pm$0.0013 & 2/4 \\
 & FSN & 99.13$\pm$0.12 & 85.36$\pm$12.44 & 94.57$\pm$2.51 & 98.67$\pm$0.33 & 0.2742$\pm$0.0049 & 0.1202$\pm$0.0016 & 0/4 \\
 & FRAPP\'E & 99.20$\pm$0.16 & 91.80$\pm$2.66 & 96.01$\pm$0.82 & 98.78$\pm$0.32 & 0.0108$\pm$0.0017 & 0.0196$\pm$0.0023 & -- \\
\multicolumn{9}{l}{\emph{Dem-aware (require group labels at inference; $^\dagger$)}} \\
 & Oracle$^\dagger$ & 99.26$\pm$0.15 & 93.94$\pm$2.02 & 96.80$\pm$0.65 & 98.83$\pm$0.33 & 0.0107$\pm$0.0023 & 0.0167$\pm$0.0019 & N/A \\
\bottomrule
\end{tabular}%
}
\end{minipage}
\end{center}

\begin{center}
\begin{minipage}{\textwidth}\centering
\captionof{table}{RFW fairness-gap results for ArcFace R50, FaceNet. Mean$\pm$std across folds. Bold marks the best label-free value within each backbone.}
\label{tab:appendix-rfw-fairness-3}
\resizebox{\textwidth}{!}{%
\scriptsize
\renewcommand{\arraystretch}{0.92}
\begin{tabular}{llcccccc}
\toprule
Backbone & Method & EO Gap 0.1\% & EO Gap 1.0\% & PE Gap 0.1\% & PE Gap 1.0\% & DP Gap 0.1\% & DP Gap 1.0\% \\
\midrule
ArcFace R50 & Baseline & 3.43$\pm$0.89 & 2.47$\pm$0.63 & \textbf{0.37$\pm$0.11} & 1.93$\pm$0.44 & 1.77$\pm$0.50 & 1.67$\pm$0.26 \\
 & FairCal & 3.10$\pm$1.25 & \textbf{2.37$\pm$0.90} & 0.40$\pm$0.14 & 1.63$\pm$0.46 & \textbf{1.52$\pm$0.58} & 1.37$\pm$0.49 \\
 & FALCON & 3.23$\pm$0.89 & 2.50$\pm$0.65 & \textbf{0.37$\pm$0.11} & 1.90$\pm$0.42 & 1.67$\pm$0.54 & 1.58$\pm$0.34 \\
 & FSN & 5.47$\pm$4.23 & 3.43$\pm$1.60 & 0.43$\pm$0.16 & 1.70$\pm$0.62 & 2.78$\pm$2.14 & 1.80$\pm$0.59 \\
 & FRAPP\'E & 4.50$\pm$1.70 & 2.63$\pm$1.36 & 0.40$\pm$0.14 & \textbf{1.30$\pm$0.51} & 2.28$\pm$0.78 & \textbf{1.32$\pm$0.59} \\
\multicolumn{8}{l}{\emph{Dem-aware (require group labels at inference; $^\dagger$)}} \\
 & Oracle$^\dagger$ & 4.07$\pm$1.97 & 2.70$\pm$1.04 & 0.43$\pm$0.16 & 1.30$\pm$0.58 & 1.95$\pm$1.06 & 1.23$\pm$0.60 \\
\bottomrule
\end{tabular}%
}
\end{minipage}
\end{center}

\paragraph{BFW.}
\begin{center}
\begin{minipage}{\textwidth}\centering
\captionof{table}{BFW accuracy and calibration results for ArcFace R100, AdaFace. Mean$\pm$std across folds. Bold marks the best label-free value within each backbone.}
\label{tab:appendix-bfw-accuracy-1}
\resizebox{\textwidth}{!}{%
\scriptsize
\renewcommand{\arraystretch}{0.92}
\begin{tabular}{llccccccc}
\toprule
Backbone & Method & AUROC (\%) & TPR 0.1\% & TPR 1.0\% & Min-grp AUROC (\%) & ECE & Brier & LD-TPR@0.1 \\
\midrule
ArcFace R100 & Baseline & 97.65$\pm$0.24 & 87.77$\pm$1.15 & 90.83$\pm$0.82 & 95.24$\pm$1.39 & 0.1105$\pm$0.0013 & 0.0721$\pm$0.0018 & -- \\
 & FairCal & 97.70$\pm$0.24 & \textbf{87.89$\pm$1.19} & 90.89$\pm$0.87 & 95.44$\pm$1.35 & 0.0064$\pm$0.0010 & \textbf{0.0260$\pm$0.0020} & 0/8 \\
 & FALCON & 97.65$\pm$0.25 & 87.83$\pm$1.14 & 90.85$\pm$0.83 & 95.24$\pm$1.40 & 0.1873$\pm$0.0020 & 0.0837$\pm$0.0013 & 8/8 \\
 & FSN & 97.57$\pm$0.31 & 87.70$\pm$1.12 & 90.77$\pm$0.82 & 95.02$\pm$1.60 & 0.1119$\pm$0.0023 & 0.0710$\pm$0.0019 & 0/8 \\
 & FRAPP\'E & 97.67$\pm$0.23 & 87.66$\pm$1.14 & 90.76$\pm$0.86 & 95.29$\pm$1.31 & \textbf{0.0061$\pm$0.0009} & 0.0263$\pm$0.0020 & -- \\
\multicolumn{9}{l}{\emph{Dem-aware (require group labels at inference; $^\dagger$)}} \\
 & Oracle$^\dagger$ & 98.41$\pm$0.17 & 87.84$\pm$1.25 & 90.94$\pm$0.87 & 96.95$\pm$0.70 & 0.0141$\pm$0.0023 & 0.0257$\pm$0.0017 & N/A \\
\bottomrule
\end{tabular}%
}
\end{minipage}
\end{center}

\begin{center}
\begin{minipage}{\textwidth}\centering
\captionof{table}{BFW fairness-gap results for ArcFace R100, AdaFace. Mean$\pm$std across folds. Bold marks the best label-free value within each backbone.}
\label{tab:appendix-bfw-fairness-1}
\resizebox{\textwidth}{!}{%
\scriptsize
\renewcommand{\arraystretch}{0.92}
\begin{tabular}{llcccccc}
\toprule
Backbone & Method & EO Gap 0.1\% & EO Gap 1.0\% & PE Gap 0.1\% & PE Gap 1.0\% & DP Gap 0.1\% & DP Gap 1.0\% \\
\midrule
ArcFace R100 & Baseline & 14.83$\pm$4.17 & 12.10$\pm$3.64 & 0.26$\pm$0.02 & 1.87$\pm$0.24 & 3.75$\pm$1.11 & 2.93$\pm$0.95 \\
 & FairCal & 15.48$\pm$4.60 & 12.23$\pm$3.79 & 0.25$\pm$0.05 & 1.57$\pm$0.24 & 3.94$\pm$1.20 & 2.73$\pm$1.11 \\
 & FALCON & 14.87$\pm$4.17 & 12.11$\pm$3.72 & 0.25$\pm$0.02 & 1.80$\pm$0.23 & 3.77$\pm$1.12 & 2.92$\pm$0.99 \\
 & FSN & 15.88$\pm$4.35 & 12.73$\pm$3.77 & 0.22$\pm$0.05 & \textbf{1.49$\pm$0.27} & 4.06$\pm$1.15 & 3.07$\pm$1.21 \\
 & FRAPP\'E & \textbf{13.84$\pm$4.08} & 11.66$\pm$3.39 & 0.33$\pm$0.09 & 2.20$\pm$0.45 & \textbf{3.46$\pm$1.05} & 2.66$\pm$0.88 \\
\multicolumn{8}{l}{\emph{Dem-aware (require group labels at inference; $^\dagger$)}} \\
 & Oracle$^\dagger$ & 16.14$\pm$4.67 & 11.77$\pm$3.81 & 0.25$\pm$0.06 & 1.83$\pm$0.34 & 4.10$\pm$1.25 & 2.58$\pm$1.08 \\
\bottomrule
\end{tabular}%
}
\end{minipage}
\end{center}

\begin{center}
\begin{minipage}{\textwidth}\centering
\captionof{table}{BFW accuracy and calibration results for MagFace. Mean$\pm$std across folds. Bold marks the best label-free value within each backbone.}
\label{tab:appendix-bfw-accuracy-2}
\resizebox{\textwidth}{!}{%
\scriptsize
\renewcommand{\arraystretch}{0.92}
\begin{tabular}{llccccccc}
\toprule
Backbone & Method & AUROC (\%) & TPR 0.1\% & TPR 1.0\% & Min-grp AUROC (\%) & ECE & Brier & LD-TPR@0.1 \\
\midrule
MagFace & Baseline & 97.05$\pm$0.39 & 83.99$\pm$1.40 & 88.87$\pm$0.99 & 94.16$\pm$1.74 & 0.1147$\pm$0.0019 & 0.0764$\pm$0.0023 & -- \\
 & FairCal & 97.22$\pm$0.39 & 84.16$\pm$1.54 & 88.96$\pm$1.08 & 94.56$\pm$1.57 & 0.0105$\pm$0.0045 & \textbf{0.0311$\pm$0.0023} & 3/8 \\
 & FALCON & 97.05$\pm$0.39 & 84.04$\pm$1.44 & 88.89$\pm$1.00 & 94.16$\pm$1.75 & 0.1801$\pm$0.0026 & 0.0891$\pm$0.0018 & 3/8 \\
 & FSN & 97.00$\pm$0.40 & 83.87$\pm$1.50 & 88.77$\pm$1.03 & 94.05$\pm$1.84 & 0.1155$\pm$0.0022 & 0.0753$\pm$0.0022 & 0/8 \\
 & FRAPP\'E & 97.09$\pm$0.34 & 83.72$\pm$1.49 & 88.86$\pm$1.05 & 94.29$\pm$1.55 & \textbf{0.0099$\pm$0.0006} & 0.0313$\pm$0.0023 & -- \\
\multicolumn{9}{l}{\emph{Dem-aware (require group labels at inference; $^\dagger$)}} \\
 & Oracle$^\dagger$ & 98.01$\pm$0.22 & 84.09$\pm$1.68 & 89.02$\pm$1.08 & 96.37$\pm$0.92 & 0.0181$\pm$0.0029 & 0.0306$\pm$0.0021 & N/A \\
 & GST$^\dagger$ & 97.01$\pm$0.40 & 84.09$\pm$1.52 & 88.84$\pm$0.98 & 94.16$\pm$1.74 & 0.0076$\pm$0.0010 & 0.0312$\pm$0.0023 & N/A \\
 & DemoNorm$^\dagger$ & 97.04$\pm$0.39 & 84.13$\pm$1.58 & 88.81$\pm$0.98 & 94.16$\pm$1.74 & 0.2270$\pm$0.0055 & 0.1922$\pm$0.0054 & N/A \\
\midrule
 & AC-MLP & 96.91$\pm$0.32 & 82.56$\pm$1.87 & 87.99$\pm$1.17 & 94.52$\pm$1.42 & 0.0237$\pm$0.0025 & 0.0341$\pm$0.0028 & 0/8 \\
 & AC-Linear & 97.35$\pm$0.37 & 83.21$\pm$1.67 & 88.87$\pm$1.02 & 94.77$\pm$1.64 & 0.0122$\pm$0.0019 & 0.0313$\pm$0.0020 & 1/8 \\
 & AC-Density & \textbf{97.44$\pm$0.26} & \textbf{84.28$\pm$1.41} & \textbf{89.08$\pm$1.00} & \textbf{94.89$\pm$1.33} & 0.0397$\pm$0.0040 & 0.0327$\pm$0.0019 & 6/8 \\
\bottomrule
\end{tabular}%
}
\end{minipage}
\end{center}

\begin{center}
\begin{minipage}{\textwidth}\centering
\captionof{table}{BFW fairness-gap results for MagFace. Mean$\pm$std across folds. Bold marks the best label-free value within each backbone.}
\label{tab:appendix-bfw-fairness-2}
\resizebox{\textwidth}{!}{%
\scriptsize
\renewcommand{\arraystretch}{0.92}
\begin{tabular}{llcccccc}
\toprule
Backbone & Method & EO Gap 0.1\% & EO Gap 1.0\% & PE Gap 0.1\% & PE Gap 1.0\% & DP Gap 0.1\% & DP Gap 1.0\% \\
\midrule
MagFace & Baseline & 17.78$\pm$3.75 & 13.64$\pm$4.12 & 0.20$\pm$0.05 & 1.25$\pm$0.20 & 4.56$\pm$1.01 & 3.26$\pm$1.34 \\
 & FairCal & 17.47$\pm$3.67 & 13.32$\pm$4.42 & 0.19$\pm$0.05 & 1.11$\pm$0.40 & 4.49$\pm$0.99 & 3.26$\pm$1.36 \\
 & FALCON & 17.86$\pm$3.75 & 13.67$\pm$4.12 & 0.20$\pm$0.05 & 1.22$\pm$0.18 & 4.58$\pm$1.01 & 3.26$\pm$1.35 \\
 & FSN & 18.75$\pm$3.08 & 14.34$\pm$4.16 & \textbf{0.18$\pm$0.06} & \textbf{0.99$\pm$0.12} & 4.83$\pm$0.85 & 3.52$\pm$1.32 \\
 & FRAPP\'E & \textbf{16.14$\pm$3.66} & 12.61$\pm$3.87 & 0.27$\pm$0.06 & 1.63$\pm$0.36 & \textbf{4.10$\pm$0.98} & 3.01$\pm$1.12 \\
\multicolumn{8}{l}{\emph{Dem-aware (require group labels at inference; $^\dagger$)}} \\
 & Oracle$^\dagger$ & 18.32$\pm$3.86 & 12.69$\pm$4.23 & 0.19$\pm$0.06 & 1.51$\pm$0.40 & 4.73$\pm$1.03 & 3.06$\pm$1.30 \\
 & GST$^\dagger$ & 18.81$\pm$3.51 & 14.22$\pm$4.24 & 0.18$\pm$0.05 & 0.85$\pm$0.22 & 4.85$\pm$0.94 & 3.53$\pm$1.33 \\
 & DemoNorm$^\dagger$ & 19.26$\pm$3.13 & 14.35$\pm$4.19 & 0.16$\pm$0.06 & 0.71$\pm$0.23 & 4.97$\pm$0.82 & 3.61$\pm$1.27 \\
\midrule
 & AC-MLP & 19.81$\pm$3.93 & 14.21$\pm$3.56 & 0.18$\pm$0.03 & 1.27$\pm$0.25 & 5.08$\pm$1.10 & 3.09$\pm$1.10 \\
 & AC-Linear & 17.06$\pm$2.82 & \textbf{12.53$\pm$3.87} & 0.21$\pm$0.03 & 1.24$\pm$0.12 & 4.33$\pm$0.80 & \textbf{2.75$\pm$1.06} \\
 & AC-Density & 17.13$\pm$3.56 & 13.18$\pm$4.16 & 0.21$\pm$0.04 & 1.27$\pm$0.20 & 4.37$\pm$0.97 & 3.11$\pm$1.30 \\
\bottomrule
\end{tabular}%
}
\end{minipage}
\end{center}

\begin{center}
\begin{minipage}{\textwidth}\centering
\captionof{table}{BFW accuracy and calibration results for ArcFace R50, FaceNet. Mean$\pm$std across folds. Bold marks the best label-free value within each backbone.}
\label{tab:appendix-bfw-accuracy-3}
\resizebox{\textwidth}{!}{%
\scriptsize
\renewcommand{\arraystretch}{0.92}
\begin{tabular}{llccccccc}
\toprule
Backbone & Method & AUROC (\%) & TPR 0.1\% & TPR 1.0\% & Min-grp AUROC (\%) & ECE & Brier & LD-TPR@0.1 \\
\midrule
ArcFace R50 & Baseline & 97.07$\pm$0.29 & 84.64$\pm$1.35 & 89.19$\pm$0.77 & 94.63$\pm$1.30 & 0.1243$\pm$0.0013 & 0.0893$\pm$0.0015 & -- \\
 & FairCal & 97.33$\pm$0.29 & \textbf{84.94$\pm$1.29} & 89.35$\pm$0.82 & 95.09$\pm$1.11 & 0.0130$\pm$0.0029 & 0.0304$\pm$0.0022 & 0/8 \\
 & FALCON & 97.07$\pm$0.29 & 84.73$\pm$1.34 & 89.23$\pm$0.76 & 94.64$\pm$1.32 & 0.2028$\pm$0.0020 & 0.0974$\pm$0.0012 & 0/8 \\
 & FSN & 96.96$\pm$0.33 & 84.65$\pm$1.25 & 89.10$\pm$0.78 & 94.32$\pm$1.51 & 0.1246$\pm$0.0017 & 0.0881$\pm$0.0014 & 0/8 \\
 & FRAPP\'E & 97.10$\pm$0.27 & 84.18$\pm$1.49 & 89.08$\pm$0.83 & 94.73$\pm$1.19 & \textbf{0.0098$\pm$0.0006} & 0.0309$\pm$0.0019 & -- \\
\multicolumn{9}{l}{\emph{Dem-aware (require group labels at inference; $^\dagger$)}} \\
 & Oracle$^\dagger$ & 97.98$\pm$0.17 & 84.79$\pm$1.33 & 89.28$\pm$0.79 & 96.42$\pm$0.83 & 0.0174$\pm$0.0020 & 0.0303$\pm$0.0016 & N/A \\
\bottomrule
\end{tabular}%
}
\end{minipage}
\end{center}

\begin{center}
\begin{minipage}{\textwidth}\centering
\captionof{table}{BFW fairness-gap results for ArcFace R50, FaceNet. Mean$\pm$std across folds. Bold marks the best label-free value within each backbone.}
\label{tab:appendix-bfw-fairness-3}
\resizebox{\textwidth}{!}{%
\scriptsize
\renewcommand{\arraystretch}{0.92}
\begin{tabular}{llcccccc}
\toprule
Backbone & Method & EO Gap 0.1\% & EO Gap 1.0\% & PE Gap 0.1\% & PE Gap 1.0\% & DP Gap 0.1\% & DP Gap 1.0\% \\
\midrule
ArcFace R50 & Baseline & 16.75$\pm$2.54 & 12.60$\pm$2.85 & 0.26$\pm$0.03 & 1.81$\pm$0.25 & 4.26$\pm$0.71 & 3.03$\pm$0.75 \\
 & FairCal & 17.44$\pm$2.52 & 12.69$\pm$3.03 & 0.24$\pm$0.06 & 1.44$\pm$0.13 & 4.41$\pm$0.66 & 3.07$\pm$0.85 \\
 & FALCON & 16.93$\pm$2.60 & 12.68$\pm$2.95 & 0.26$\pm$0.02 & 1.72$\pm$0.23 & 4.31$\pm$0.73 & 3.04$\pm$0.84 \\
 & FSN & 18.63$\pm$2.63 & 13.75$\pm$3.29 & 0.20$\pm$0.05 & \textbf{1.28$\pm$0.40} & 4.80$\pm$0.73 & 3.28$\pm$1.16 \\
 & FRAPP\'E & \textbf{15.53$\pm$2.47} & 12.32$\pm$2.68 & 0.36$\pm$0.06 & 2.38$\pm$0.35 & \textbf{3.92$\pm$0.68} & 2.80$\pm$0.68 \\
\multicolumn{8}{l}{\emph{Dem-aware (require group labels at inference; $^\dagger$)}} \\
 & Oracle$^\dagger$ & 18.60$\pm$2.93 & 12.59$\pm$2.99 & 0.23$\pm$0.04 & 1.78$\pm$0.15 & 4.74$\pm$0.77 & 2.98$\pm$0.89 \\
\bottomrule
\end{tabular}%
}
\end{minipage}
\end{center}

\paragraph{LFW.}
\begin{center}
\begin{minipage}{\textwidth}\centering
\captionof{table}{LFW accuracy and calibration results for ArcFace R100, AdaFace. Mean$\pm$std across folds. Bold marks the best label-free value within each backbone.}
\label{tab:appendix-lfw-accuracy-1}
\resizebox{\textwidth}{!}{%
\scriptsize
\renewcommand{\arraystretch}{0.92}
\begin{tabular}{llccccccc}
\toprule
Backbone & Method & AUROC (\%) & TPR 0.1\% & TPR 1.0\% & Min-grp AUROC (\%) & ECE & Brier & LD-TPR@0.1 \\
\midrule
ArcFace R100 & Baseline & 97.12$\pm$0.75 & 59.77$\pm$12.52 & 72.30$\pm$11.28 & 91.60$\pm$6.22 & 0.2257$\pm$0.0089 & 0.1494$\pm$0.0077 & -- \\
 & FairCal & 97.86$\pm$0.72 & 60.40$\pm$16.32 & 76.67$\pm$6.65 & 93.36$\pm$5.13 & 0.0343$\pm$0.0065 & 0.0539$\pm$0.0087 & 0/8 \\
 & FALCON & 97.29$\pm$0.72 & \textbf{63.77$\pm$13.69} & 75.93$\pm$8.88 & 91.71$\pm$6.29 & 0.2804$\pm$0.0111 & 0.1923$\pm$0.0065 & 0/8 \\
 & FSN & 95.34$\pm$1.13 & 48.70$\pm$12.10 & 63.47$\pm$8.11 & 90.55$\pm$5.89 & 0.1889$\pm$0.0128 & 0.1208$\pm$0.0054 & 0/8 \\
 & FRAPP\'E & 96.51$\pm$0.99 & 57.07$\pm$11.62 & 73.57$\pm$8.32 & 90.45$\pm$5.16 & \textbf{0.0322$\pm$0.0063} & 0.0666$\pm$0.0100 & -- \\
\multicolumn{9}{l}{\emph{Dem-aware (require group labels at inference; $^\dagger$)}} \\
 & Oracle$^\dagger$ & 97.17$\pm$0.71 & 56.33$\pm$15.96 & 72.60$\pm$11.01 & 91.82$\pm$6.27 & 0.0344$\pm$0.0068 & 0.0599$\pm$0.0090 & N/A \\
\bottomrule
\end{tabular}%
}
\end{minipage}
\end{center}

\begin{center}
\begin{minipage}{\textwidth}\centering
\captionof{table}{LFW fairness-gap results for ArcFace R100, AdaFace. Mean$\pm$std across folds. Bold marks the best label-free value within each backbone.}
\label{tab:appendix-lfw-fairness-1}
\resizebox{\textwidth}{!}{%
\scriptsize
\renewcommand{\arraystretch}{0.92}
\begin{tabular}{llcccccc}
\toprule
Backbone & Method & EO Gap 0.1\% & EO Gap 1.0\% & PE Gap 0.1\% & PE Gap 1.0\% & DP Gap 0.1\% & DP Gap 1.0\% \\
\midrule
ArcFace R100 & Baseline & 46.09$\pm$26.65 & 41.33$\pm$25.77 & 1.65$\pm$1.50 & 8.98$\pm$14.92 & 40.12$\pm$13.66 & 46.96$\pm$19.96 \\
 & FairCal & 52.28$\pm$21.27 & 45.03$\pm$22.77 & 1.34$\pm$0.70 & 4.68$\pm$3.77 & 39.78$\pm$15.58 & 47.44$\pm$15.05 \\
 & FALCON & 46.39$\pm$25.43 & 39.91$\pm$25.08 & \textbf{1.25$\pm$0.59} & 8.83$\pm$14.98 & 41.34$\pm$15.11 & 47.35$\pm$19.58 \\
 & FSN & 48.63$\pm$20.32 & 50.44$\pm$22.20 & 3.58$\pm$5.04 & 5.73$\pm$4.79 & 35.99$\pm$9.98 & 42.00$\pm$12.08 \\
 & FRAPP\'E & 50.67$\pm$23.30 & 44.84$\pm$24.77 & 2.14$\pm$1.62 & 4.93$\pm$3.67 & 39.26$\pm$16.93 & 43.61$\pm$17.34 \\
\multicolumn{8}{l}{\emph{Dem-aware (require group labels at inference; $^\dagger$)}} \\
 & Oracle$^\dagger$ & 41.33$\pm$18.83 & 41.52$\pm$24.69 & 2.33$\pm$2.58 & 8.87$\pm$14.94 & 37.98$\pm$10.81 & 47.54$\pm$18.23 \\
\bottomrule
\end{tabular}%
}
\end{minipage}
\end{center}

\begin{center}
\begin{minipage}{\textwidth}\centering
\captionof{table}{LFW accuracy and calibration results for MagFace. Mean$\pm$std across folds. Bold marks the best label-free value within each backbone.}
\label{tab:appendix-lfw-accuracy-2}
\resizebox{\textwidth}{!}{%
\scriptsize
\renewcommand{\arraystretch}{0.92}
\begin{tabular}{llccccccc}
\toprule
Backbone & Method & AUROC (\%) & TPR 0.1\% & TPR 1.0\% & Min-grp AUROC (\%) & ECE & Brier & LD-TPR@0.1 \\
\midrule
MagFace & Baseline & 95.47$\pm$1.15 & 46.17$\pm$16.04 & 66.80$\pm$8.96 & 84.79$\pm$9.97 & 0.2196$\pm$0.0141 & 0.1640$\pm$0.0078 & -- \\
 & FairCal & 96.58$\pm$0.88 & 49.30$\pm$15.92 & 70.43$\pm$7.53 & 88.77$\pm$6.94 & \textbf{0.0357$\pm$0.0045} & 0.0706$\pm$0.0101 & 0/8 \\
 & FALCON & 95.82$\pm$1.10 & 50.33$\pm$15.46 & 69.53$\pm$7.59 & 88.34$\pm$6.74 & 0.2759$\pm$0.0151 & 0.2131$\pm$0.0064 & 0/8 \\
 & FSN & 94.68$\pm$1.42 & 38.73$\pm$18.40 & 63.10$\pm$9.77 & 87.20$\pm$7.52 & 0.2007$\pm$0.0161 & 0.1327$\pm$0.0062 & 0/8 \\
 & FRAPP\'E & 94.39$\pm$1.14 & 43.77$\pm$12.40 & 61.50$\pm$9.26 & 76.44$\pm$27.56 & 0.0383$\pm$0.0097 & 0.0894$\pm$0.0104 & -- \\
\multicolumn{9}{l}{\emph{Dem-aware (require group labels at inference; $^\dagger$)}} \\
 & Oracle$^\dagger$ & 95.62$\pm$0.99 & 46.27$\pm$17.45 & 65.73$\pm$7.57 & 84.88$\pm$10.20 & 0.0398$\pm$0.0078 & 0.0779$\pm$0.0104 & N/A \\
 & GST$^\dagger$ & 95.42$\pm$1.20 & 47.27$\pm$16.85 & 66.50$\pm$7.44 & 84.79$\pm$9.97 & 0.0392$\pm$0.0064 & 0.0795$\pm$0.0116 & N/A \\
 & DemoNorm$^\dagger$ & 95.46$\pm$1.23 & 47.77$\pm$17.50 & 65.47$\pm$8.27 & 84.79$\pm$9.97 & 0.1540$\pm$0.0163 & 0.1438$\pm$0.0175 & N/A \\
\midrule
 & AC-MLP & 92.45$\pm$1.24 & 25.23$\pm$15.94 & 42.13$\pm$15.21 & 85.17$\pm$5.17 & 0.1616$\pm$0.0135 & 0.1375$\pm$0.0056 & 0/8 \\
 & AC-Linear & \textbf{97.18$\pm$0.77} & \textbf{50.80$\pm$20.99} & \textbf{73.37$\pm$8.53} & \textbf{90.57$\pm$8.76} & 0.0477$\pm$0.0068 & \textbf{0.0634$\pm$0.0093} & 0/8 \\
 & AC-Density & 96.73$\pm$0.98 & 44.70$\pm$17.00 & 68.10$\pm$15.38 & 90.52$\pm$6.29 & 0.0574$\pm$0.0103 & 0.0705$\pm$0.0087 & 0/8 \\
\bottomrule
\end{tabular}%
}
\end{minipage}
\end{center}

\begin{center}
\begin{minipage}{\textwidth}\centering
\captionof{table}{LFW fairness-gap results for MagFace. Mean$\pm$std across folds. Bold marks the best label-free value within each backbone.}
\label{tab:appendix-lfw-fairness-2}
\resizebox{\textwidth}{!}{%
\scriptsize
\renewcommand{\arraystretch}{0.92}
\begin{tabular}{llcccccc}
\toprule
Backbone & Method & EO Gap 0.1\% & EO Gap 1.0\% & PE Gap 0.1\% & PE Gap 1.0\% & DP Gap 0.1\% & DP Gap 1.0\% \\
\midrule
MagFace & Baseline & 44.52$\pm$26.21 & 52.50$\pm$22.69 & 3.81$\pm$2.97 & 5.30$\pm$3.44 & 33.84$\pm$13.57 & 45.88$\pm$14.20 \\
 & FairCal & 44.51$\pm$25.18 & 46.10$\pm$27.07 & 3.73$\pm$2.97 & \textbf{4.40$\pm$2.64} & 30.50$\pm$13.42 & 42.66$\pm$15.98 \\
 & FALCON & 43.35$\pm$26.32 & 47.33$\pm$23.47 & 3.73$\pm$2.97 & 4.76$\pm$2.75 & 34.29$\pm$14.07 & 43.05$\pm$13.27 \\
 & FSN & 40.84$\pm$20.82 & 53.42$\pm$22.39 & 2.27$\pm$1.83 & 4.73$\pm$2.71 & 24.12$\pm$12.85 & 38.59$\pm$11.49 \\
 & FRAPP\'E & 56.63$\pm$19.19 & 54.64$\pm$24.96 & 3.30$\pm$3.28 & 4.91$\pm$3.05 & 34.87$\pm$11.91 & 42.91$\pm$16.03 \\
\multicolumn{8}{l}{\emph{Dem-aware (require group labels at inference; $^\dagger$)}} \\
 & Oracle$^\dagger$ & 42.98$\pm$23.53 & 49.89$\pm$25.57 & 3.62$\pm$2.47 & 4.57$\pm$2.63 & 30.55$\pm$12.01 & 42.28$\pm$12.56 \\
 & GST$^\dagger$ & 43.82$\pm$23.55 & 50.89$\pm$25.86 & 3.22$\pm$2.44 & 4.32$\pm$1.88 & 32.87$\pm$14.64 & 43.81$\pm$16.17 \\
 & DemoNorm$^\dagger$ & 47.42$\pm$23.16 & 50.22$\pm$26.49 & 3.10$\pm$2.45 & 4.86$\pm$2.39 & 32.95$\pm$15.04 & 41.09$\pm$16.89 \\
\midrule
 & AC-MLP & \textbf{31.28$\pm$15.16} & 45.59$\pm$26.86 & \textbf{1.72$\pm$1.60} & 4.82$\pm$4.55 & \textbf{18.10$\pm$8.81} & \textbf{28.60$\pm$7.02} \\
 & AC-Linear & 49.10$\pm$26.47 & 44.72$\pm$29.38 & 3.02$\pm$2.75 & 5.47$\pm$4.37 & 31.27$\pm$13.58 & 42.67$\pm$15.54 \\
 & AC-Density & 46.11$\pm$25.02 & \textbf{40.06$\pm$25.97} & 2.34$\pm$1.97 & 5.02$\pm$4.36 & 28.47$\pm$14.22 & 39.33$\pm$16.14 \\
\bottomrule
\end{tabular}%
}
\end{minipage}
\end{center}

\begin{center}
\begin{minipage}{\textwidth}\centering
\captionof{table}{LFW accuracy and calibration results for ArcFace R50, ArcFace (FairFace). Mean$\pm$std across folds. Bold marks the best label-free value within each backbone.}
\label{tab:appendix-lfw-accuracy-3}
\resizebox{\textwidth}{!}{%
\scriptsize
\renewcommand{\arraystretch}{0.92}
\begin{tabular}{llccccccc}
\toprule
Backbone & Method & AUROC (\%) & TPR 0.1\% & TPR 1.0\% & Min-grp AUROC (\%) & ECE & Brier & LD-TPR@0.1 \\
\midrule
ArcFace R50 & Baseline & 98.53$\pm$0.40 & \textbf{96.30$\pm$0.58} & 96.47$\pm$0.63 & 91.77$\pm$9.02 & 0.1707$\pm$0.0036 & 0.0757$\pm$0.0038 & -- \\
 & FairCal & 98.46$\pm$0.47 & 96.20$\pm$0.50 & \textbf{96.50$\pm$0.69} & 92.18$\pm$3.93 & 0.0123$\pm$0.0049 & 0.0190$\pm$0.0029 & 0/1 \\
 & FALCON & 98.53$\pm$0.40 & \textbf{96.30$\pm$0.58} & 96.47$\pm$0.63 & 91.77$\pm$9.03 & 0.2297$\pm$0.0046 & 0.0870$\pm$0.0026 & 0/1 \\
 & FSN & 98.30$\pm$0.43 & 96.00$\pm$0.70 & 96.23$\pm$0.69 & 91.04$\pm$7.94 & 0.1549$\pm$0.0068 & 0.0519$\pm$0.0031 & 0/1 \\
 & FRAPP\'E & 98.52$\pm$0.41 & \textbf{96.30$\pm$0.58} & \textbf{96.50$\pm$0.69} & 91.52$\pm$9.09 & \textbf{0.0081$\pm$0.0022} & \textbf{0.0183$\pm$0.0029} & -- \\
\multicolumn{9}{l}{\emph{Dem-aware (require group labels at inference; $^\dagger$)}} \\
 & Oracle$^\dagger$ & 98.53$\pm$0.40 & 96.30$\pm$0.58 & 96.47$\pm$0.63 & 91.77$\pm$9.02 & 0.0085$\pm$0.0023 & 0.0183$\pm$0.0028 & N/A \\
\bottomrule
\end{tabular}%
}
\end{minipage}
\end{center}

\begin{center}
\begin{minipage}{\textwidth}\centering
\captionof{table}{LFW fairness-gap results for ArcFace R50, ArcFace (FairFace). Mean$\pm$std across folds. Bold marks the best label-free value within each backbone.}
\label{tab:appendix-lfw-fairness-3}
\resizebox{\textwidth}{!}{%
\scriptsize
\renewcommand{\arraystretch}{0.92}
\begin{tabular}{llcccccc}
\toprule
Backbone & Method & EO Gap 0.1\% & EO Gap 1.0\% & PE Gap 0.1\% & PE Gap 1.0\% & DP Gap 0.1\% & DP Gap 1.0\% \\
\midrule
ArcFace R50 & Baseline & 14.18$\pm$7.18 & \textbf{13.57$\pm$7.52} & 1.43$\pm$0.71 & \textbf{3.66$\pm$2.46} & 48.42$\pm$15.92 & 48.52$\pm$16.43 \\
 & FairCal & \textbf{13.82$\pm$7.50} & \textbf{13.57$\pm$7.52} & 1.41$\pm$0.62 & 4.00$\pm$2.38 & 48.37$\pm$16.01 & 48.59$\pm$16.17 \\
 & FALCON & 14.18$\pm$7.18 & \textbf{13.57$\pm$7.52} & 1.43$\pm$0.71 & \textbf{3.66$\pm$2.46} & 48.42$\pm$15.92 & 48.52$\pm$16.43 \\
 & FSN & 15.01$\pm$7.95 & 13.94$\pm$7.13 & 1.57$\pm$0.73 & 3.96$\pm$3.37 & 48.21$\pm$16.22 & 48.16$\pm$16.32 \\
 & FRAPP\'E & 14.18$\pm$7.18 & \textbf{13.57$\pm$7.52} & 1.43$\pm$0.71 & \textbf{3.66$\pm$2.46} & 48.42$\pm$15.92 & 48.52$\pm$16.43 \\
\multicolumn{8}{l}{\emph{Dem-aware (require group labels at inference; $^\dagger$)}} \\
 & Oracle$^\dagger$ & 14.18$\pm$7.18 & 13.57$\pm$7.52 & 1.43$\pm$0.71 & 3.66$\pm$2.46 & 48.42$\pm$15.92 & 48.52$\pm$16.43 \\
\bottomrule
\end{tabular}%
}
\end{minipage}
\end{center}

\begin{center}
\begin{minipage}{\textwidth}\centering
\captionof{table}{LFW accuracy and calibration results for FaceNet. Mean$\pm$std across folds. Bold marks the best label-free value within each backbone.}
\label{tab:appendix-lfw-accuracy-4}
\resizebox{\textwidth}{!}{%
\scriptsize
\renewcommand{\arraystretch}{0.92}
\begin{tabular}{llccccccc}
\toprule
Backbone & Method & AUROC (\%) & TPR 0.1\% & TPR 1.0\% & Min-grp AUROC (\%) & ECE & Brier & LD-TPR@0.1 \\
\midrule
FaceNet & Baseline & 99.27$\pm$0.34 & 81.40$\pm$14.03 & 91.87$\pm$2.82 & 94.65$\pm$8.32 & 0.1710$\pm$0.0099 & 0.0743$\pm$0.0057 & -- \\
 & FairCal & 99.10$\pm$0.36 & 65.40$\pm$28.97 & 91.00$\pm$3.22 & 96.56$\pm$4.55 & \textbf{0.0204$\pm$0.0044} & 0.0258$\pm$0.0048 & 0/8 \\
 & FALCON & 99.29$\pm$0.34 & \textbf{82.80$\pm$12.58} & 92.40$\pm$2.78 & 94.68$\pm$8.34 & 0.2555$\pm$0.0072 & 0.1475$\pm$0.0057 & 7/8 \\
 & FSN & 95.58$\pm$1.48 & 32.37$\pm$10.82 & 43.43$\pm$10.54 & 85.36$\pm$10.88 & 0.1328$\pm$0.0065 & 0.1103$\pm$0.0143 & 0/8 \\
 & FRAPP\'E & 98.73$\pm$0.57 & 63.90$\pm$17.06 & 83.73$\pm$6.63 & 93.93$\pm$5.30 & 0.0242$\pm$0.0041 & 0.0382$\pm$0.0098 & -- \\
\multicolumn{9}{l}{\emph{Dem-aware (require group labels at inference; $^\dagger$)}} \\
 & Oracle$^\dagger$ & 99.22$\pm$0.29 & 81.23$\pm$15.54 & 91.13$\pm$3.41 & 93.51$\pm$8.32 & 0.0219$\pm$0.0050 & 0.0253$\pm$0.0054 & N/A \\
 & GST$^\dagger$ & 99.21$\pm$0.43 & 81.40$\pm$14.03 & 92.27$\pm$2.80 & 94.65$\pm$8.32 & 0.0221$\pm$0.0050 & 0.0262$\pm$0.0059 & N/A \\
 & DemoNorm$^\dagger$ & 99.28$\pm$0.35 & 82.30$\pm$13.54 & 92.53$\pm$2.27 & 94.65$\pm$8.32 & 0.1566$\pm$0.0118 & 0.1320$\pm$0.0121 & N/A \\
\midrule
 & AC-MLP & 98.03$\pm$0.85 & 59.83$\pm$10.37 & 70.53$\pm$8.96 & 94.03$\pm$5.31 & 0.1500$\pm$0.0152 & 0.0780$\pm$0.0074 & 0/8 \\
 & AC-Linear & 99.39$\pm$0.26 & 80.77$\pm$17.29 & \textbf{93.10$\pm$2.66} & 97.06$\pm$4.67 & 0.0285$\pm$0.0029 & \textbf{0.0233$\pm$0.0046} & 7/8 \\
 & AC-Density & \textbf{99.42$\pm$0.23} & 76.47$\pm$21.05 & 92.63$\pm$2.58 & \textbf{97.21$\pm$4.69} & 0.0328$\pm$0.0036 & 0.0248$\pm$0.0050 & 1/8 \\
\bottomrule
\end{tabular}%
}
\end{minipage}
\end{center}

\begin{center}
\begin{minipage}{\textwidth}\centering
\captionof{table}{LFW fairness-gap results for FaceNet. Mean$\pm$std across folds. Bold marks the best label-free value within each backbone.}
\label{tab:appendix-lfw-fairness-4}
\resizebox{\textwidth}{!}{%
\scriptsize
\renewcommand{\arraystretch}{0.92}
\begin{tabular}{llcccccc}
\toprule
Backbone & Method & EO Gap 0.1\% & EO Gap 1.0\% & PE Gap 0.1\% & PE Gap 1.0\% & DP Gap 0.1\% & DP Gap 1.0\% \\
\midrule
FaceNet & Baseline & 29.18$\pm$13.55 & 24.67$\pm$11.14 & \textbf{1.25$\pm$0.56} & 4.32$\pm$5.55 & 43.62$\pm$16.12 & 46.53$\pm$17.60 \\
 & FairCal & 30.18$\pm$13.48 & 20.05$\pm$8.07 & 1.89$\pm$2.32 & \textbf{3.47$\pm$2.35} & 38.86$\pm$18.43 & 49.54$\pm$17.22 \\
 & FALCON & 24.71$\pm$12.14 & 22.73$\pm$9.53 & \textbf{1.25$\pm$0.56} & 4.45$\pm$5.53 & 43.35$\pm$17.94 & 46.74$\pm$18.06 \\
 & FSN & 41.22$\pm$9.88 & 50.57$\pm$9.98 & 4.19$\pm$6.61 & 6.24$\pm$5.97 & \textbf{22.73$\pm$13.30} & \textbf{30.85$\pm$16.86} \\
 & FRAPP\'E & 43.92$\pm$8.00 & 31.83$\pm$11.07 & 1.78$\pm$1.19 & 4.30$\pm$3.70 & 38.77$\pm$18.08 & 46.23$\pm$15.98 \\
\multicolumn{8}{l}{\emph{Dem-aware (require group labels at inference; $^\dagger$)}} \\
 & Oracle$^\dagger$ & 29.71$\pm$13.73 & 23.60$\pm$11.24 & 1.92$\pm$2.10 & 5.45$\pm$5.51 & 43.67$\pm$16.82 & 48.51$\pm$17.23 \\
 & GST$^\dagger$ & 53.33$\pm$19.93 & 40.49$\pm$27.92 & 1.49$\pm$1.14 & 4.09$\pm$2.36 & 48.62$\pm$15.15 & 54.01$\pm$13.10 \\
 & DemoNorm$^\dagger$ & 26.32$\pm$11.42 & 21.42$\pm$9.81 & 1.59$\pm$1.12 & 4.57$\pm$5.50 & 44.22$\pm$18.31 & 48.63$\pm$17.90 \\
\midrule
 & AC-MLP & 39.49$\pm$18.22 & 30.35$\pm$16.83 & 2.17$\pm$1.69 & 3.86$\pm$2.04 & 38.28$\pm$15.00 & 41.82$\pm$14.41 \\
 & AC-Linear & \textbf{23.19$\pm$13.78} & \textbf{18.83$\pm$7.60} & 3.26$\pm$5.91 & 4.77$\pm$5.50 & 45.36$\pm$16.95 & 48.53$\pm$18.70 \\
 & AC-Density & 27.20$\pm$12.51 & 21.85$\pm$10.18 & \textbf{1.25$\pm$0.56} & 4.32$\pm$5.55 & 45.04$\pm$17.91 & 47.67$\pm$18.44 \\
\bottomrule
\end{tabular}%
}
\end{minipage}
\end{center}

\subsubsection{One-table summaries by dataset and backbone}
\begin{center}
\begin{minipage}{\textwidth}\centering
\captionof{table}{Results for ArcFace R100 on RFW.}
\label{tab:appendix-summary-rfw-arcface_r100}
\resizebox{\textwidth}{!}{%
\footnotesize
\renewcommand{\arraystretch}{1.2}
\setlength{\tabcolsep}{3.5pt}
\begin{tabular}{l c c c c c c c c c}
\toprule
\textbf{Method} & \textbf{AUROC (\%) $\uparrow$} & \textbf{TPR (0.1\%) $\uparrow$} & \textbf{TPR (1.0\%) $\uparrow$} & \textbf{PE Gap (0.1\%) $\downarrow$} & \textbf{PE Gap (1.0\%) $\downarrow$} & \textbf{ECE $\downarrow$} & \textbf{Brier $\downarrow$} & \textbf{Min-grp AUROC (\%) $\uparrow$} & \textbf{LD-TPR@0.1} $\uparrow$ \\
\midrule
Baseline & 97.42$\pm$0.14 & 61.19$\pm$5.68 & 77.61$\pm$2.50 & 0.47$\pm$0.17 & 1.80$\pm$0.65 & 0.2677$\pm$0.0039 & 0.1675$\pm$0.0017 & 96.27$\pm$0.38 & -- \\
FairCal & 97.83$\pm$0.18 & \textbf{65.30$\pm$3.90} & 80.07$\pm$2.87 & 0.43$\pm$0.16 & 1.57$\pm$0.45 & \textbf{0.0168$\pm$0.0030} & 0.0526$\pm$0.0018 & 96.69$\pm$0.36 & 0/4 \\
FALCON & 97.46$\pm$0.14 & 61.96$\pm$5.49 & 78.06$\pm$2.51 & 0.50$\pm$0.18 & 1.80$\pm$0.55 & 0.3044$\pm$0.0039 & 0.1786$\pm$0.0013 & 96.30$\pm$0.39 & 4/4 \\
FSN & 97.16$\pm$0.19 & 57.64$\pm$7.68 & 76.38$\pm$2.85 & 0.40$\pm$0.14 & \textbf{1.40$\pm$0.56} & 0.2596$\pm$0.0037 & 0.1531$\pm$0.0018 & 95.82$\pm$0.37 & 0/4 \\
FRAPP\'E & 97.33$\pm$0.14 & 61.06$\pm$4.72 & 77.02$\pm$2.13 & 0.40$\pm$0.14 & 1.83$\pm$0.48 & 0.0190$\pm$0.0039 & 0.0592$\pm$0.0019 & 95.93$\pm$0.61 & 2/4 \\
\midrule
\multicolumn{10}{l}{\emph{Dem-aware (require group labels at inference)}} \\
Oracle$^\dagger$ & 97.58$\pm$0.16 & 62.94$\pm$5.63 & 79.06$\pm$2.58 & 0.43$\pm$0.16 & 1.47$\pm$0.50 & 0.0164$\pm$0.0033 & 0.0559$\pm$0.0014 & 96.27$\pm$0.38 & N/A \\
\bottomrule
\end{tabular}%
}
\end{minipage}
\end{center}

\begin{center}
\begin{minipage}{\textwidth}\centering
\captionof{table}{Results for AdaFace on RFW.}
\label{tab:appendix-summary-rfw-adaface}
\resizebox{\textwidth}{!}{%
\footnotesize
\renewcommand{\arraystretch}{1.2}
\setlength{\tabcolsep}{3.5pt}
\begin{tabular}{l c c c c c c c c c}
\toprule
\textbf{Method} & \textbf{AUROC (\%) $\uparrow$} & \textbf{TPR (0.1\%) $\uparrow$} & \textbf{TPR (1.0\%) $\uparrow$} & \textbf{PE Gap (0.1\%) $\downarrow$} & \textbf{PE Gap (1.0\%) $\downarrow$} & \textbf{ECE $\downarrow$} & \textbf{Brier $\downarrow$} & \textbf{Min-grp AUROC (\%) $\uparrow$} & \textbf{LD-TPR@0.1} $\uparrow$ \\
\midrule
Baseline & 98.39$\pm$0.23 & 69.97$\pm$6.79 & 83.39$\pm$2.09 & 0.47$\pm$0.17 & 1.80$\pm$0.23 & 0.2762$\pm$0.0033 & 0.1700$\pm$0.0016 & 97.60$\pm$0.43 & -- \\
FairCal & 98.63$\pm$0.25 & \textbf{70.83$\pm$6.90} & 85.58$\pm$1.86 & \textbf{0.37$\pm$0.11} & 1.53$\pm$0.32 & 0.0137$\pm$0.0023 & 0.0414$\pm$0.0029 & 97.73$\pm$0.49 & 1/4 \\
FALCON & 98.41$\pm$0.23 & 70.37$\pm$6.68 & 83.52$\pm$2.03 & 0.47$\pm$0.17 & 1.83$\pm$0.24 & 0.3093$\pm$0.0039 & 0.1701$\pm$0.0012 & 97.60$\pm$0.43 & 4/4 \\
FSN & 98.12$\pm$0.30 & 65.43$\pm$6.48 & 82.26$\pm$2.82 & \textbf{0.37$\pm$0.11} & 1.83$\pm$0.36 & 0.2678$\pm$0.0055 & 0.1550$\pm$0.0014 & 97.14$\pm$0.51 & 1/4 \\
FRAPP\'E & 98.43$\pm$0.23 & 70.49$\pm$6.00 & 83.91$\pm$1.86 & \textbf{0.37$\pm$0.11} & 1.77$\pm$0.45 & \textbf{0.0123$\pm$0.0016} & 0.0439$\pm$0.0030 & 97.51$\pm$0.38 & 2/4 \\
\midrule
\multicolumn{10}{l}{\emph{Dem-aware (require group labels at inference)}} \\
Oracle$^\dagger$ & 98.52$\pm$0.25 & 73.13$\pm$4.89 & 85.13$\pm$1.65 & 0.37$\pm$0.11 & 1.43$\pm$0.45 & 0.0121$\pm$0.0026 & 0.0425$\pm$0.0032 & 97.60$\pm$0.43 & N/A \\
\bottomrule
\end{tabular}%
}
\end{minipage}
\end{center}

\begin{center}
\begin{minipage}{\textwidth}\centering
\captionof{table}{Results for MagFace on RFW.}
\label{tab:appendix-summary-rfw-magface}
\resizebox{\textwidth}{!}{%
\footnotesize
\renewcommand{\arraystretch}{1.2}
\setlength{\tabcolsep}{3.5pt}
\begin{tabular}{l c c c c c c c c c}
\toprule
\textbf{Method} & \textbf{AUROC (\%) $\uparrow$} & \textbf{TPR (0.1\%) $\uparrow$} & \textbf{TPR (1.0\%) $\uparrow$} & \textbf{PE Gap (0.1\%) $\downarrow$} & \textbf{PE Gap (1.0\%) $\downarrow$} & \textbf{ECE $\downarrow$} & \textbf{Brier $\downarrow$} & \textbf{Min-grp AUROC (\%) $\uparrow$} & \textbf{LD-TPR@0.1} $\uparrow$ \\
\midrule
Baseline & 95.63$\pm$0.36 & 48.81$\pm$8.04 & 66.20$\pm$3.92 & 0.53$\pm$0.17 & 1.70$\pm$0.43 & 0.2440$\pm$0.0042 & 0.1713$\pm$0.0025 & 94.08$\pm$0.82 & -- \\
FairCal & 96.13$\pm$0.34 & 51.24$\pm$5.98 & 68.92$\pm$3.18 & 0.43$\pm$0.16 & \textbf{1.30$\pm$0.58} & \textbf{0.0194$\pm$0.0037} & 0.0737$\pm$0.0036 & 94.49$\pm$0.79 & 1/4 \\
FALCON & 95.69$\pm$0.36 & 50.02$\pm$8.27 & 66.68$\pm$3.90 & 0.53$\pm$0.17 & 1.57$\pm$0.57 & 0.2785$\pm$0.0065 & 0.1882$\pm$0.0020 & 94.13$\pm$0.83 & 4/4 \\
FSN & 95.31$\pm$0.39 & 49.56$\pm$7.84 & 65.73$\pm$4.52 & \textbf{0.40$\pm$0.14} & 2.07$\pm$0.58 & 0.2390$\pm$0.0062 & 0.1597$\pm$0.0025 & 93.28$\pm$1.02 & 1/4 \\
FRAPP\'E & 95.70$\pm$0.41 & \textbf{51.55$\pm$6.99} & 66.40$\pm$2.94 & 0.43$\pm$0.16 & 1.40$\pm$0.68 & 0.0216$\pm$0.0043 & 0.0771$\pm$0.0044 & 93.86$\pm$0.94 & 3/4 \\
\midrule
\multicolumn{10}{l}{\emph{Dem-aware (require group labels at inference)}} \\
Oracle$^\dagger$ & 95.85$\pm$0.36 & 51.04$\pm$7.73 & 68.42$\pm$3.19 & 0.47$\pm$0.17 & 1.20$\pm$0.72 & 0.0217$\pm$0.0039 & 0.0759$\pm$0.0037 & 94.08$\pm$0.82 & N/A \\
\bottomrule
\end{tabular}%
}
\end{minipage}
\end{center}

\begin{center}
\begin{minipage}{\textwidth}\centering
\captionof{table}{Results for ArcFace R50 on RFW.}
\label{tab:appendix-summary-rfw-arcface}
\resizebox{\textwidth}{!}{%
\footnotesize
\renewcommand{\arraystretch}{1.2}
\setlength{\tabcolsep}{3.5pt}
\begin{tabular}{l c c c c c c c c c}
\toprule
\textbf{Method} & \textbf{AUROC (\%) $\uparrow$} & \textbf{TPR (0.1\%) $\uparrow$} & \textbf{TPR (1.0\%) $\uparrow$} & \textbf{PE Gap (0.1\%) $\downarrow$} & \textbf{PE Gap (1.0\%) $\downarrow$} & \textbf{ECE $\downarrow$} & \textbf{Brier $\downarrow$} & \textbf{Min-grp AUROC (\%) $\uparrow$} & \textbf{LD-TPR@0.1} $\uparrow$ \\
\midrule
Baseline & 99.25$\pm$0.15 & 93.14$\pm$2.51 & 96.58$\pm$0.77 & \textbf{0.37$\pm$0.11} & 1.93$\pm$0.44 & 0.2797$\pm$0.0031 & 0.1286$\pm$0.0018 & 98.83$\pm$0.33 & -- \\
FairCal & 99.26$\pm$0.15 & 93.49$\pm$1.58 & 96.77$\pm$0.73 & 0.40$\pm$0.14 & 1.63$\pm$0.46 & \textbf{0.0102$\pm$0.0017} & 0.0170$\pm$0.0020 & 98.81$\pm$0.33 & 3/4 \\
FALCON & 99.25$\pm$0.15 & 93.14$\pm$2.64 & 96.64$\pm$0.78 & \textbf{0.37$\pm$0.11} & 1.90$\pm$0.42 & 0.3236$\pm$0.0034 & 0.1406$\pm$0.0013 & 98.83$\pm$0.33 & 2/4 \\
FSN & 99.13$\pm$0.12 & 85.36$\pm$12.44 & 94.57$\pm$2.51 & 0.43$\pm$0.16 & 1.70$\pm$0.62 & 0.2742$\pm$0.0049 & 0.1202$\pm$0.0016 & 98.67$\pm$0.33 & 0/4 \\
FRAPP\'E & 99.20$\pm$0.16 & 91.80$\pm$2.66 & 96.01$\pm$0.82 & 0.40$\pm$0.14 & \textbf{1.30$\pm$0.51} & 0.0108$\pm$0.0017 & 0.0196$\pm$0.0023 & 98.78$\pm$0.32 & 0/4 \\
\midrule
\multicolumn{10}{l}{\emph{Dem-aware (require group labels at inference)}} \\
Oracle$^\dagger$ & 99.26$\pm$0.15 & 93.94$\pm$2.02 & 96.80$\pm$0.65 & 0.43$\pm$0.16 & 1.30$\pm$0.58 & 0.0107$\pm$0.0023 & 0.0167$\pm$0.0019 & 98.83$\pm$0.33 & N/A \\
\bottomrule
\end{tabular}%
}
\end{minipage}
\end{center}

\begin{center}
\begin{minipage}{\textwidth}\centering
\captionof{table}{Results for FaceNet on RFW.}
\label{tab:appendix-summary-rfw-facenet}
\resizebox{\textwidth}{!}{%
\footnotesize
\renewcommand{\arraystretch}{1.2}
\setlength{\tabcolsep}{3.5pt}
\begin{tabular}{l c c c c c c c c c}
\toprule
\textbf{Method} & \textbf{AUROC (\%) $\uparrow$} & \textbf{TPR (0.1\%) $\uparrow$} & \textbf{TPR (1.0\%) $\uparrow$} & \textbf{PE Gap (0.1\%) $\downarrow$} & \textbf{PE Gap (1.0\%) $\downarrow$} & \textbf{ECE $\downarrow$} & \textbf{Brier $\downarrow$} & \textbf{Min-grp AUROC (\%) $\uparrow$} & \textbf{LD-TPR@0.1} $\uparrow$ \\
\midrule
Baseline & 87.83$\pm$0.44 & 14.99$\pm$2.52 & 30.12$\pm$3.01 & 0.53$\pm$0.17 & 2.13$\pm$0.59 & 0.1213$\pm$0.0059 & 0.1607$\pm$0.0017 & 86.14$\pm$1.02 & -- \\
FairCal & 90.57$\pm$0.50 & 19.52$\pm$5.49 & 39.73$\pm$3.79 & \textbf{0.37$\pm$0.11} & \textbf{1.13$\pm$0.61} & 0.0233$\pm$0.0053 & 0.1236$\pm$0.0038 & 87.81$\pm$1.18 & 3/4 \\
FALCON & 88.84$\pm$0.47 & 17.92$\pm$2.90 & 34.18$\pm$2.86 & 0.50$\pm$0.18 & 1.90$\pm$0.59 & 0.1911$\pm$0.0056 & 0.2000$\pm$0.0021 & 86.69$\pm$1.06 & 3/4 \\
FSN & 89.57$\pm$0.62 & 19.58$\pm$4.08 & 36.95$\pm$3.64 & \textbf{0.37$\pm$0.11} & 1.23$\pm$0.52 & 0.1596$\pm$0.0065 & 0.1620$\pm$0.0022 & 86.51$\pm$1.24 & 3/4 \\
FRAPP\'E & 89.57$\pm$0.64 & 19.07$\pm$4.63 & 37.40$\pm$3.70 & \textbf{0.37$\pm$0.11} & 1.33$\pm$0.63 & \textbf{0.0207$\pm$0.0033} & 0.1303$\pm$0.0044 & 86.39$\pm$1.34 & 1/4 \\
\midrule
\multicolumn{10}{l}{\emph{Dem-aware (require group labels at inference)}} \\
Oracle$^\dagger$ & 89.56$\pm$0.54 & 19.56$\pm$4.34 & 37.80$\pm$3.32 & 0.50$\pm$0.18 & 1.17$\pm$0.59 & 0.0238$\pm$0.0044 & 0.1304$\pm$0.0038 & 86.14$\pm$1.02 & N/A \\
\bottomrule
\end{tabular}%
}
\end{minipage}
\end{center}

\begin{center}
\begin{minipage}{\textwidth}\centering
\captionof{table}{Results for ArcFace R100 on BFW.}
\label{tab:appendix-summary-bfw-arcface_r100}
\resizebox{\textwidth}{!}{%
\footnotesize
\renewcommand{\arraystretch}{1.2}
\setlength{\tabcolsep}{3.5pt}
\begin{tabular}{l c c c c c c c c c}
\toprule
\textbf{Method} & \textbf{AUROC (\%) $\uparrow$} & \textbf{TPR (0.1\%) $\uparrow$} & \textbf{TPR (1.0\%) $\uparrow$} & \textbf{PE Gap (0.1\%) $\downarrow$} & \textbf{PE Gap (1.0\%) $\downarrow$} & \textbf{ECE $\downarrow$} & \textbf{Brier $\downarrow$} & \textbf{Min-grp AUROC (\%) $\uparrow$} & \textbf{LD-TPR@0.1} $\uparrow$ \\
\midrule
Baseline & 97.65$\pm$0.24 & 87.77$\pm$1.15 & 90.83$\pm$0.82 & 0.26$\pm$0.02 & 1.87$\pm$0.24 & 0.1105$\pm$0.0013 & 0.0721$\pm$0.0018 & 95.24$\pm$1.39 & -- \\
FairCal & 97.70$\pm$0.24 & \textbf{87.89$\pm$1.19} & 90.89$\pm$0.87 & 0.25$\pm$0.05 & 1.57$\pm$0.24 & 0.0064$\pm$0.0010 & \textbf{0.0260$\pm$0.0020} & 95.44$\pm$1.35 & 0/8 \\
FALCON & 97.65$\pm$0.25 & 87.83$\pm$1.14 & 90.85$\pm$0.83 & 0.25$\pm$0.02 & 1.80$\pm$0.23 & 0.1873$\pm$0.0020 & 0.0837$\pm$0.0013 & 95.24$\pm$1.40 & 8/8 \\
FSN & 97.57$\pm$0.31 & 87.70$\pm$1.12 & 90.77$\pm$0.82 & 0.22$\pm$0.05 & \textbf{1.49$\pm$0.27} & 0.1119$\pm$0.0023 & 0.0710$\pm$0.0019 & 95.02$\pm$1.60 & 0/8 \\
FRAPP\'E & 97.67$\pm$0.23 & 87.66$\pm$1.14 & 90.76$\pm$0.86 & 0.33$\pm$0.09 & 2.20$\pm$0.45 & \textbf{0.0061$\pm$0.0009} & 0.0263$\pm$0.0020 & 95.29$\pm$1.31 & 3/8 \\
\midrule
\multicolumn{10}{l}{\emph{Dem-aware (require group labels at inference)}} \\
Oracle$^\dagger$ & 98.41$\pm$0.17 & 87.84$\pm$1.25 & 90.94$\pm$0.87 & 0.25$\pm$0.06 & 1.83$\pm$0.34 & 0.0141$\pm$0.0023 & 0.0257$\pm$0.0017 & 96.95$\pm$0.70 & N/A \\
\bottomrule
\end{tabular}%
}
\end{minipage}
\end{center}

\begin{center}
\begin{minipage}{\textwidth}\centering
\captionof{table}{Results for AdaFace on BFW.}
\label{tab:appendix-summary-bfw-adaface}
\resizebox{\textwidth}{!}{%
\footnotesize
\renewcommand{\arraystretch}{1.2}
\setlength{\tabcolsep}{3.5pt}
\begin{tabular}{l c c c c c c c c c}
\toprule
\textbf{Method} & \textbf{AUROC (\%) $\uparrow$} & \textbf{TPR (0.1\%) $\uparrow$} & \textbf{TPR (1.0\%) $\uparrow$} & \textbf{PE Gap (0.1\%) $\downarrow$} & \textbf{PE Gap (1.0\%) $\downarrow$} & \textbf{ECE $\downarrow$} & \textbf{Brier $\downarrow$} & \textbf{Min-grp AUROC (\%) $\uparrow$} & \textbf{LD-TPR@0.1} $\uparrow$ \\
\midrule
Baseline & 97.28$\pm$0.33 & 87.28$\pm$1.19 & 90.54$\pm$0.80 & 0.29$\pm$0.02 & 1.92$\pm$0.24 & 0.1184$\pm$0.0010 & 0.0814$\pm$0.0018 & 94.82$\pm$1.37 & -- \\
FairCal & 97.24$\pm$0.41 & 87.45$\pm$1.18 & 90.64$\pm$0.83 & 0.28$\pm$0.06 & 1.54$\pm$0.35 & 0.0196$\pm$0.0085 & 0.0282$\pm$0.0028 & 94.64$\pm$1.42 & 0/8 \\
FALCON & 97.28$\pm$0.34 & 87.31$\pm$1.18 & 90.57$\pm$0.80 & 0.28$\pm$0.02 & 1.83$\pm$0.21 & 0.1983$\pm$0.0021 & 0.0885$\pm$0.0013 & 94.82$\pm$1.39 & 0/8 \\
FSN & 97.17$\pm$0.39 & 87.28$\pm$1.12 & 90.49$\pm$0.85 & \textbf{0.20$\pm$0.03} & \textbf{1.30$\pm$0.33} & 0.1167$\pm$0.0011 & 0.0795$\pm$0.0019 & 94.67$\pm$1.64 & 0/8 \\
FRAPP\'E & 97.30$\pm$0.34 & 86.93$\pm$1.21 & 90.41$\pm$0.82 & 0.37$\pm$0.05 & 2.57$\pm$0.32 & \textbf{0.0068$\pm$0.0009} & \textbf{0.0274$\pm$0.0020} & 94.95$\pm$1.27 & 1/8 \\
\midrule
\multicolumn{10}{l}{\emph{Dem-aware (require group labels at inference)}} \\
Oracle$^\dagger$ & 98.19$\pm$0.22 & 87.40$\pm$1.22 & 90.66$\pm$0.82 & 0.25$\pm$0.08 & 1.95$\pm$0.15 & 0.0156$\pm$0.0023 & 0.0268$\pm$0.0018 & 96.65$\pm$0.82 & N/A \\
\bottomrule
\end{tabular}%
}
\end{minipage}
\end{center}

\begin{center}
\begin{minipage}{\textwidth}\centering
\captionof{table}{Results for MagFace on BFW.}
\label{tab:appendix-summary-bfw-magface}
\resizebox{\textwidth}{!}{%
\footnotesize
\renewcommand{\arraystretch}{1.2}
\setlength{\tabcolsep}{3.5pt}
\begin{tabular}{l c c c c c c c c c}
\toprule
\textbf{Method} & \textbf{AUROC (\%) $\uparrow$} & \textbf{TPR (0.1\%) $\uparrow$} & \textbf{TPR (1.0\%) $\uparrow$} & \textbf{PE Gap (0.1\%) $\downarrow$} & \textbf{PE Gap (1.0\%) $\downarrow$} & \textbf{ECE $\downarrow$} & \textbf{Brier $\downarrow$} & \textbf{Min-grp AUROC (\%) $\uparrow$} & \textbf{LD-TPR@0.1} $\uparrow$ \\
\midrule
Baseline & 97.05$\pm$0.39 & 83.99$\pm$1.40 & 88.87$\pm$0.99 & 0.20$\pm$0.05 & 1.25$\pm$0.20 & 0.1147$\pm$0.0019 & 0.0764$\pm$0.0023 & 94.16$\pm$1.74 & -- \\
FairCal & 97.22$\pm$0.39 & 84.16$\pm$1.54 & 88.96$\pm$1.08 & 0.19$\pm$0.05 & 1.11$\pm$0.40 & 0.0105$\pm$0.0045 & \textbf{0.0311$\pm$0.0023} & 94.56$\pm$1.57 & 3/8 \\
FALCON & 97.05$\pm$0.39 & 84.04$\pm$1.44 & 88.89$\pm$1.00 & 0.20$\pm$0.05 & 1.22$\pm$0.18 & 0.1801$\pm$0.0026 & 0.0891$\pm$0.0018 & 94.16$\pm$1.75 & 3/8 \\
FSN & 97.00$\pm$0.40 & 83.87$\pm$1.50 & 88.77$\pm$1.03 & \textbf{0.18$\pm$0.06} & \textbf{0.99$\pm$0.12} & 0.1155$\pm$0.0022 & 0.0753$\pm$0.0022 & 94.05$\pm$1.84 & 0/8 \\
FRAPP\'E & 97.09$\pm$0.34 & 83.72$\pm$1.49 & 88.86$\pm$1.05 & 0.27$\pm$0.06 & 1.63$\pm$0.36 & \textbf{0.0099$\pm$0.0006} & 0.0313$\pm$0.0023 & 94.29$\pm$1.55 & 2/8 \\
\midrule
\multicolumn{10}{l}{\emph{Dem-aware (require group labels at inference)}} \\
Oracle$^\dagger$ & 98.01$\pm$0.22 & 84.09$\pm$1.68 & 89.02$\pm$1.08 & 0.19$\pm$0.06 & 1.51$\pm$0.40 & 0.0181$\pm$0.0029 & 0.0306$\pm$0.0021 & 96.37$\pm$0.92 & N/A \\
\bottomrule
\end{tabular}%
}
\end{minipage}
\end{center}

\begin{center}
\begin{minipage}{\textwidth}\centering
\captionof{table}{Results for ArcFace R50 on BFW.}
\label{tab:appendix-summary-bfw-arcface}
\resizebox{\textwidth}{!}{%
\footnotesize
\renewcommand{\arraystretch}{1.2}
\setlength{\tabcolsep}{3.5pt}
\begin{tabular}{l c c c c c c c c c}
\toprule
\textbf{Method} & \textbf{AUROC (\%) $\uparrow$} & \textbf{TPR (0.1\%) $\uparrow$} & \textbf{TPR (1.0\%) $\uparrow$} & \textbf{PE Gap (0.1\%) $\downarrow$} & \textbf{PE Gap (1.0\%) $\downarrow$} & \textbf{ECE $\downarrow$} & \textbf{Brier $\downarrow$} & \textbf{Min-grp AUROC (\%) $\uparrow$} & \textbf{LD-TPR@0.1} $\uparrow$ \\
\midrule
Baseline & 97.07$\pm$0.29 & 84.64$\pm$1.35 & 89.19$\pm$0.77 & 0.26$\pm$0.03 & 1.81$\pm$0.25 & 0.1243$\pm$0.0013 & 0.0893$\pm$0.0015 & 94.63$\pm$1.30 & -- \\
FairCal & 97.33$\pm$0.29 & \textbf{84.94$\pm$1.29} & 89.35$\pm$0.82 & 0.24$\pm$0.06 & 1.44$\pm$0.13 & 0.0130$\pm$0.0029 & 0.0304$\pm$0.0022 & 95.09$\pm$1.11 & 0/8 \\
FALCON & 97.07$\pm$0.29 & 84.73$\pm$1.34 & 89.23$\pm$0.76 & 0.26$\pm$0.02 & 1.72$\pm$0.23 & 0.2028$\pm$0.0020 & 0.0974$\pm$0.0012 & 94.64$\pm$1.32 & 0/8 \\
FSN & 96.96$\pm$0.33 & 84.65$\pm$1.25 & 89.10$\pm$0.78 & 0.20$\pm$0.05 & \textbf{1.28$\pm$0.40} & 0.1246$\pm$0.0017 & 0.0881$\pm$0.0014 & 94.32$\pm$1.51 & 0/8 \\
FRAPP\'E & 97.10$\pm$0.27 & 84.18$\pm$1.49 & 89.08$\pm$0.83 & 0.36$\pm$0.06 & 2.38$\pm$0.35 & \textbf{0.0098$\pm$0.0006} & 0.0309$\pm$0.0019 & 94.73$\pm$1.19 & 1/8 \\
\midrule
\multicolumn{10}{l}{\emph{Dem-aware (require group labels at inference)}} \\
Oracle$^\dagger$ & 97.98$\pm$0.17 & 84.79$\pm$1.33 & 89.28$\pm$0.79 & 0.23$\pm$0.04 & 1.78$\pm$0.15 & 0.0174$\pm$0.0020 & 0.0303$\pm$0.0016 & 96.42$\pm$0.83 & N/A \\
\bottomrule
\end{tabular}%
}
\end{minipage}
\end{center}

\begin{center}
\begin{minipage}{\textwidth}\centering
\captionof{table}{Results for FaceNet on BFW.}
\label{tab:appendix-summary-bfw-facenet}
\resizebox{\textwidth}{!}{%
\footnotesize
\renewcommand{\arraystretch}{1.2}
\setlength{\tabcolsep}{3.5pt}
\begin{tabular}{l c c c c c c c c c}
\toprule
\textbf{Method} & \textbf{AUROC (\%) $\uparrow$} & \textbf{TPR (0.1\%) $\uparrow$} & \textbf{TPR (1.0\%) $\uparrow$} & \textbf{PE Gap (0.1\%) $\downarrow$} & \textbf{PE Gap (1.0\%) $\downarrow$} & \textbf{ECE $\downarrow$} & \textbf{Brier $\downarrow$} & \textbf{Min-grp AUROC (\%) $\uparrow$} & \textbf{LD-TPR@0.1} $\uparrow$ \\
\midrule
Baseline & 98.80$\pm$0.13 & 67.95$\pm$2.75 & 85.70$\pm$1.49 & 0.30$\pm$0.10 & 2.22$\pm$0.40 & 0.1194$\pm$0.0039 & 0.0561$\pm$0.0009 & \textbf{97.78$\pm$0.45} & -- \\
FairCal & 98.90$\pm$0.12 & 73.15$\pm$2.28 & \textbf{87.78$\pm$1.23} & 0.21$\pm$0.07 & 1.57$\pm$0.37 & 0.0039$\pm$0.0005 & \textbf{0.0287$\pm$0.0018} & 97.77$\pm$0.47 & 1/8 \\
FALCON & 98.85$\pm$0.13 & 70.37$\pm$2.44 & 86.62$\pm$1.36 & 0.26$\pm$0.08 & 1.99$\pm$0.39 & 0.1404$\pm$0.0024 & 0.0826$\pm$0.0008 & 97.77$\pm$0.47 & 1/8 \\
FSN & 98.85$\pm$0.12 & 71.82$\pm$1.85 & 87.12$\pm$1.10 & \textbf{0.17$\pm$0.04} & \textbf{1.47$\pm$0.35} & 0.1305$\pm$0.0027 & 0.0569$\pm$0.0008 & 97.68$\pm$0.50 & 1/8 \\
FRAPP\'E & 98.86$\pm$0.13 & 71.26$\pm$2.71 & 87.13$\pm$1.34 & 0.23$\pm$0.09 & 1.65$\pm$0.38 & \textbf{0.0033$\pm$0.0005} & 0.0295$\pm$0.0019 & 97.72$\pm$0.49 & 1/8 \\
\midrule
\multicolumn{10}{l}{\emph{Dem-aware (require group labels at inference)}} \\
Oracle$^\dagger$ & 98.95$\pm$0.13 & 72.27$\pm$2.22 & 87.43$\pm$1.25 & 0.16$\pm$0.07 & 1.37$\pm$0.48 & 0.0060$\pm$0.0027 & 0.0291$\pm$0.0017 & 97.88$\pm$0.47 & N/A \\
\bottomrule
\end{tabular}%
}
\end{minipage}
\end{center}

\begin{center}
\begin{minipage}{\textwidth}\centering
\captionof{table}{Results for ArcFace R100 on LFW.}
\label{tab:appendix-summary-lfw-arcface_r100}
\resizebox{\textwidth}{!}{%
\footnotesize
\renewcommand{\arraystretch}{1.2}
\setlength{\tabcolsep}{3.5pt}
\begin{tabular}{l c c c c c c c c c}
\toprule
\textbf{Method} & \textbf{AUROC (\%) $\uparrow$} & \textbf{TPR (0.1\%) $\uparrow$} & \textbf{TPR (1.0\%) $\uparrow$} & \textbf{PE Gap (0.1\%) $\downarrow$} & \textbf{PE Gap (1.0\%) $\downarrow$} & \textbf{ECE $\downarrow$} & \textbf{Brier $\downarrow$} & \textbf{Min-grp AUROC (\%) $\uparrow$} & \textbf{LD-TPR@0.1} $\uparrow$ \\
\midrule
Baseline & 97.12$\pm$0.75 & 59.77$\pm$12.52 & 72.30$\pm$11.28 & 1.65$\pm$1.50 & 8.98$\pm$14.92 & 0.2257$\pm$0.0089 & 0.1494$\pm$0.0077 & 91.60$\pm$6.22 & -- \\
FairCal & 97.86$\pm$0.72 & 60.40$\pm$16.32 & 76.67$\pm$6.65 & 1.34$\pm$0.70 & 4.68$\pm$3.77 & 0.0343$\pm$0.0065 & 0.0539$\pm$0.0087 & 93.36$\pm$5.13 & 0/8 \\
FALCON & 97.29$\pm$0.72 & \textbf{63.77$\pm$13.69} & 75.93$\pm$8.88 & \textbf{1.25$\pm$0.59} & 8.83$\pm$14.98 & 0.2804$\pm$0.0111 & 0.1923$\pm$0.0065 & 91.71$\pm$6.29 & 0/8 \\
FSN & 95.34$\pm$1.13 & 48.70$\pm$12.10 & 63.47$\pm$8.11 & 3.58$\pm$5.04 & 5.73$\pm$4.79 & 0.1889$\pm$0.0128 & 0.1208$\pm$0.0054 & 90.55$\pm$5.89 & 0/8 \\
FRAPP\'E & 96.51$\pm$0.99 & 57.07$\pm$11.62 & 73.57$\pm$8.32 & 2.14$\pm$1.62 & 4.93$\pm$3.67 & \textbf{0.0322$\pm$0.0063} & 0.0666$\pm$0.0100 & 90.45$\pm$5.16 & 0/8 \\
\midrule
\multicolumn{10}{l}{\emph{Dem-aware (require group labels at inference)}} \\
Oracle$^\dagger$ & 97.17$\pm$0.71 & 56.33$\pm$15.96 & 72.60$\pm$11.01 & 2.33$\pm$2.58 & 8.87$\pm$14.94 & 0.0344$\pm$0.0068 & 0.0599$\pm$0.0090 & 91.82$\pm$6.27 & N/A \\
\bottomrule
\end{tabular}%
}
\end{minipage}
\end{center}

\begin{center}
\begin{minipage}{\textwidth}\centering
\captionof{table}{Results for AdaFace on LFW.}
\label{tab:appendix-summary-lfw-adaface}
\resizebox{\textwidth}{!}{%
\footnotesize
\renewcommand{\arraystretch}{1.2}
\setlength{\tabcolsep}{3.5pt}
\begin{tabular}{l c c c c c c c c c}
\toprule
\textbf{Method} & \textbf{AUROC (\%) $\uparrow$} & \textbf{TPR (0.1\%) $\uparrow$} & \textbf{TPR (1.0\%) $\uparrow$} & \textbf{PE Gap (0.1\%) $\downarrow$} & \textbf{PE Gap (1.0\%) $\downarrow$} & \textbf{ECE $\downarrow$} & \textbf{Brier $\downarrow$} & \textbf{Min-grp AUROC (\%) $\uparrow$} & \textbf{LD-TPR@0.1} $\uparrow$ \\
\midrule
Baseline & 98.96$\pm$0.41 & 87.00$\pm$6.73 & 93.27$\pm$2.72 & 2.14$\pm$1.47 & 5.08$\pm$4.47 & 0.2572$\pm$0.0076 & 0.1582$\pm$0.0070 & 97.25$\pm$2.12 & -- \\
FairCal & 99.32$\pm$0.32 & 83.47$\pm$8.14 & 92.67$\pm$2.02 & 2.33$\pm$1.40 & 4.78$\pm$3.61 & \textbf{0.0204$\pm$0.0040} & \textbf{0.0256$\pm$0.0058} & 97.59$\pm$2.64 & 1/8 \\
FALCON & 98.92$\pm$0.40 & \textbf{87.30$\pm$6.20} & 93.20$\pm$2.64 & 2.14$\pm$1.47 & 5.08$\pm$4.47 & 0.3003$\pm$0.0088 & 0.1688$\pm$0.0053 & 97.12$\pm$2.07 & 1/8 \\
FSN & 96.70$\pm$0.90 & 65.70$\pm$14.31 & 77.37$\pm$5.19 & 3.24$\pm$5.95 & 4.06$\pm$5.63 & 0.1906$\pm$0.0137 & 0.1168$\pm$0.0044 & 91.90$\pm$3.45 & 0/8 \\
FRAPP\'E & 98.84$\pm$0.39 & 84.97$\pm$7.49 & 91.50$\pm$1.71 & 2.12$\pm$1.93 & \textbf{3.01$\pm$1.49} & 0.0210$\pm$0.0051 & 0.0318$\pm$0.0065 & 96.96$\pm$2.00 & 2/8 \\
\midrule
\multicolumn{10}{l}{\emph{Dem-aware (require group labels at inference)}} \\
Oracle$^\dagger$ & 98.96$\pm$0.38 & 85.83$\pm$5.96 & 93.13$\pm$2.79 & 2.94$\pm$2.36 & 5.13$\pm$3.05 & 0.0217$\pm$0.0047 & 0.0282$\pm$0.0065 & 97.41$\pm$1.98 & N/A \\
\bottomrule
\end{tabular}%
}
\end{minipage}
\end{center}

\begin{center}
\begin{minipage}{\textwidth}\centering
\captionof{table}{Results for MagFace on LFW.}
\label{tab:appendix-summary-lfw-magface}
\resizebox{\textwidth}{!}{%
\footnotesize
\renewcommand{\arraystretch}{1.2}
\setlength{\tabcolsep}{3.5pt}
\begin{tabular}{l c c c c c c c c c}
\toprule
\textbf{Method} & \textbf{AUROC (\%) $\uparrow$} & \textbf{TPR (0.1\%) $\uparrow$} & \textbf{TPR (1.0\%) $\uparrow$} & \textbf{PE Gap (0.1\%) $\downarrow$} & \textbf{PE Gap (1.0\%) $\downarrow$} & \textbf{ECE $\downarrow$} & \textbf{Brier $\downarrow$} & \textbf{Min-grp AUROC (\%) $\uparrow$} & \textbf{LD-TPR@0.1} $\uparrow$ \\
\midrule
Baseline & 95.47$\pm$1.15 & 46.17$\pm$16.04 & 66.80$\pm$8.96 & 3.81$\pm$2.97 & 5.30$\pm$3.44 & 0.2196$\pm$0.0141 & 0.1640$\pm$0.0078 & 84.79$\pm$9.97 & -- \\
FairCal & 96.58$\pm$0.88 & 49.30$\pm$15.92 & 70.43$\pm$7.53 & 3.73$\pm$2.97 & \textbf{4.40$\pm$2.64} & \textbf{0.0357$\pm$0.0045} & 0.0706$\pm$0.0101 & 88.77$\pm$6.94 & 0/8 \\
FALCON & 95.82$\pm$1.10 & 50.33$\pm$15.46 & 69.53$\pm$7.59 & 3.73$\pm$2.97 & 4.76$\pm$2.75 & 0.2759$\pm$0.0151 & 0.2131$\pm$0.0064 & 88.34$\pm$6.74 & 0/8 \\
FSN & 94.68$\pm$1.42 & 38.73$\pm$18.40 & 63.10$\pm$9.77 & 2.27$\pm$1.83 & 4.73$\pm$2.71 & 0.2007$\pm$0.0161 & 0.1327$\pm$0.0062 & 87.20$\pm$7.52 & 0/8 \\
FRAPP\'E & 94.39$\pm$1.14 & 43.77$\pm$12.40 & 61.50$\pm$9.26 & 3.30$\pm$3.28 & 4.91$\pm$3.05 & 0.0383$\pm$0.0097 & 0.0894$\pm$0.0104 & 76.44$\pm$27.56 & 3/8 \\
\midrule
\multicolumn{10}{l}{\emph{Dem-aware (require group labels at inference)}} \\
Oracle$^\dagger$ & 95.62$\pm$0.99 & 46.27$\pm$17.45 & 65.73$\pm$7.57 & 3.62$\pm$2.47 & 4.57$\pm$2.63 & 0.0398$\pm$0.0078 & 0.0779$\pm$0.0104 & 84.88$\pm$10.20 & N/A \\
\bottomrule
\end{tabular}%
}
\end{minipage}
\end{center}

\begin{center}
\begin{minipage}{\textwidth}\centering
\captionof{table}{Results for ArcFace R50 on LFW.}
\label{tab:appendix-summary-lfw-arcface}
\resizebox{\textwidth}{!}{%
\footnotesize
\renewcommand{\arraystretch}{1.2}
\setlength{\tabcolsep}{3.5pt}
\begin{tabular}{l c c c c c c c c c}
\toprule
\textbf{Method} & \textbf{AUROC (\%) $\uparrow$} & \textbf{TPR (0.1\%) $\uparrow$} & \textbf{TPR (1.0\%) $\uparrow$} & \textbf{PE Gap (0.1\%) $\downarrow$} & \textbf{PE Gap (1.0\%) $\downarrow$} & \textbf{ECE $\downarrow$} & \textbf{Brier $\downarrow$} & \textbf{Min-grp AUROC (\%) $\uparrow$} & \textbf{LD-TPR@0.1} $\uparrow$ \\
\midrule
Baseline & 98.53$\pm$0.40 & \textbf{96.30$\pm$0.58} & 96.47$\pm$0.63 & 1.43$\pm$0.71 & \textbf{3.66$\pm$2.46} & 0.1707$\pm$0.0036 & 0.0757$\pm$0.0038 & 91.77$\pm$9.02 & -- \\
FairCal & 98.46$\pm$0.47 & 96.20$\pm$0.50 & \textbf{96.50$\pm$0.69} & 1.41$\pm$0.62 & 4.00$\pm$2.38 & 0.0123$\pm$0.0049 & 0.0190$\pm$0.0029 & 92.18$\pm$3.93 & 0/1 \\
FALCON & 98.53$\pm$0.40 & \textbf{96.30$\pm$0.58} & 96.47$\pm$0.63 & 1.43$\pm$0.71 & \textbf{3.66$\pm$2.46} & 0.2297$\pm$0.0046 & 0.0870$\pm$0.0026 & 91.77$\pm$9.03 & 0/1 \\
FSN & 98.30$\pm$0.43 & 96.00$\pm$0.70 & 96.23$\pm$0.69 & 1.57$\pm$0.73 & 3.96$\pm$3.37 & 0.1549$\pm$0.0068 & 0.0519$\pm$0.0031 & 91.04$\pm$7.94 & 0/1 \\
FRAPP\'E & 98.52$\pm$0.41 & \textbf{96.30$\pm$0.58} & \textbf{96.50$\pm$0.69} & 1.43$\pm$0.71 & \textbf{3.66$\pm$2.46} & \textbf{0.0081$\pm$0.0022} & \textbf{0.0183$\pm$0.0029} & 91.52$\pm$9.09 & 1/1 \\
\midrule
\multicolumn{10}{l}{\emph{Dem-aware (require group labels at inference)}} \\
Oracle$^\dagger$ & 98.53$\pm$0.40 & 96.30$\pm$0.58 & 96.47$\pm$0.63 & 1.43$\pm$0.71 & 3.66$\pm$2.46 & 0.0085$\pm$0.0023 & 0.0183$\pm$0.0028 & 91.77$\pm$9.02 & N/A \\
\bottomrule
\end{tabular}%
}
\end{minipage}
\end{center}

\begin{center}
\begin{minipage}{\textwidth}\centering
\captionof{table}{Results for ArcFace (FairFace) on LFW.}
\label{tab:appendix-summary-lfw-arcface_fairface}
\resizebox{\textwidth}{!}{%
\footnotesize
\renewcommand{\arraystretch}{1.2}
\setlength{\tabcolsep}{3.5pt}
\begin{tabular}{l c c c c c c c c c}
\toprule
\textbf{Method} & \textbf{AUROC (\%) $\uparrow$} & \textbf{TPR (0.1\%) $\uparrow$} & \textbf{TPR (1.0\%) $\uparrow$} & \textbf{PE Gap (0.1\%) $\downarrow$} & \textbf{PE Gap (1.0\%) $\downarrow$} & \textbf{ECE $\downarrow$} & \textbf{Brier $\downarrow$} & \textbf{Min-grp AUROC (\%) $\uparrow$} & \textbf{LD-TPR@0.1} $\uparrow$ \\
\midrule
Baseline & 98.53$\pm$0.40 & \textbf{96.30$\pm$0.58} & 96.47$\pm$0.63 & 1.43$\pm$0.71 & \textbf{3.66$\pm$2.46} & 0.1707$\pm$0.0036 & 0.0757$\pm$0.0038 & 91.77$\pm$9.02 & -- \\
FairCal & 98.46$\pm$0.47 & 96.20$\pm$0.50 & \textbf{96.50$\pm$0.69} & \textbf{1.41$\pm$0.62} & 4.00$\pm$2.38 & \textbf{0.0123$\pm$0.0049} & 0.0190$\pm$0.0029 & 92.18$\pm$3.93 & 1/8 \\
FALCON & 98.53$\pm$0.40 & \textbf{96.30$\pm$0.58} & 96.47$\pm$0.63 & 1.43$\pm$0.71 & \textbf{3.66$\pm$2.46} & 0.2297$\pm$0.0046 & 0.0870$\pm$0.0026 & 91.77$\pm$9.03 & 0/8 \\
FSN & 98.30$\pm$0.43 & 96.00$\pm$0.70 & 96.23$\pm$0.69 & 1.57$\pm$0.73 & 3.96$\pm$3.37 & 0.1549$\pm$0.0068 & 0.0519$\pm$0.0031 & 91.04$\pm$7.94 & 0/8 \\
FRAPP\'E & 98.46$\pm$0.35 & 93.83$\pm$2.84 & 95.77$\pm$0.93 & 2.86$\pm$2.72 & 5.31$\pm$3.47 & 0.0190$\pm$0.0031 & 0.0234$\pm$0.0037 & 93.62$\pm$3.24 & 0/8 \\
\midrule
\multicolumn{10}{l}{\emph{Dem-aware (require group labels at inference)}} \\
Oracle$^\dagger$ & 98.57$\pm$0.36 & 93.37$\pm$9.28 & 96.57$\pm$0.42 & 4.46$\pm$3.00 & 7.75$\pm$4.97 & 0.0100$\pm$0.0028 & 0.0185$\pm$0.0028 & 91.63$\pm$8.94 & N/A \\
\bottomrule
\end{tabular}%
}
\end{minipage}
\end{center}

\begin{center}
\begin{minipage}{\textwidth}\centering
\captionof{table}{Results for FaceNet on LFW.}
\label{tab:appendix-summary-lfw-facenet}
\resizebox{\textwidth}{!}{%
\footnotesize
\renewcommand{\arraystretch}{1.2}
\setlength{\tabcolsep}{3.5pt}
\begin{tabular}{l c c c c c c c c c}
\toprule
\textbf{Method} & \textbf{AUROC (\%) $\uparrow$} & \textbf{TPR (0.1\%) $\uparrow$} & \textbf{TPR (1.0\%) $\uparrow$} & \textbf{PE Gap (0.1\%) $\downarrow$} & \textbf{PE Gap (1.0\%) $\downarrow$} & \textbf{ECE $\downarrow$} & \textbf{Brier $\downarrow$} & \textbf{Min-grp AUROC (\%) $\uparrow$} & \textbf{LD-TPR@0.1} $\uparrow$ \\
\midrule
Baseline & 99.27$\pm$0.34 & 81.40$\pm$14.03 & 91.87$\pm$2.82 & \textbf{1.25$\pm$0.56} & 4.32$\pm$5.55 & 0.1710$\pm$0.0099 & 0.0743$\pm$0.0057 & 94.65$\pm$8.32 & -- \\
FairCal & 99.10$\pm$0.36 & 65.40$\pm$28.97 & 91.00$\pm$3.22 & 1.89$\pm$2.32 & \textbf{3.47$\pm$2.35} & \textbf{0.0204$\pm$0.0044} & 0.0258$\pm$0.0048 & 96.56$\pm$4.55 & 0/8 \\
FALCON & 99.29$\pm$0.34 & \textbf{82.80$\pm$12.58} & 92.40$\pm$2.78 & \textbf{1.25$\pm$0.56} & 4.45$\pm$5.53 & 0.2555$\pm$0.0072 & 0.1475$\pm$0.0057 & 94.68$\pm$8.34 & 7/8 \\
FSN & 95.58$\pm$1.48 & 32.37$\pm$10.82 & 43.43$\pm$10.54 & 4.19$\pm$6.61 & 6.24$\pm$5.97 & 0.1328$\pm$0.0065 & 0.1103$\pm$0.0143 & 85.36$\pm$10.88 & 0/8 \\
FRAPP\'E & 98.73$\pm$0.57 & 63.90$\pm$17.06 & 83.73$\pm$6.63 & 1.78$\pm$1.19 & 4.30$\pm$3.70 & 0.0242$\pm$0.0041 & 0.0382$\pm$0.0098 & 93.93$\pm$5.30 & 0/8 \\
\midrule
\multicolumn{10}{l}{\emph{Dem-aware (require group labels at inference)}} \\
Oracle$^\dagger$ & 99.22$\pm$0.29 & 81.23$\pm$15.54 & 91.13$\pm$3.41 & 1.92$\pm$2.10 & 5.45$\pm$5.51 & 0.0219$\pm$0.0050 & 0.0253$\pm$0.0054 & 93.51$\pm$8.32 & N/A \\
\bottomrule
\end{tabular}%
}
\end{minipage}
\end{center}

\subsubsection{Per-group AUROC}
\label{pg_AUROC}
\begin{center}
\begin{minipage}{\textwidth}\centering
\captionof{table}{Per-group AUROC for ArcFace R100 on RFW.}
\label{tab:appendix-auroc-rfw-arcface_r100}
{%
\footnotesize
\renewcommand{\arraystretch}{1.2}
\setlength{\tabcolsep}{3.5pt}
%
%
}
\end{minipage}
\end{center}

\begin{center}
\begin{minipage}{\textwidth}\centering
\captionof{table}{Per-group AUROC for AdaFace on RFW.}
\label{tab:appendix-auroc-rfw-adaface}
{%
\footnotesize
\renewcommand{\arraystretch}{1.2}
\setlength{\tabcolsep}{3.5pt}
%
%
}
\end{minipage}
\end{center}

\begin{center}
\begin{minipage}{\textwidth}\centering
\captionof{table}{Per-group AUROC for MagFace on RFW.}
\label{tab:appendix-auroc-rfw-magface}
{%
\footnotesize
\renewcommand{\arraystretch}{1.2}
\setlength{\tabcolsep}{3.5pt}
%
%
}
\end{minipage}
\end{center}

\begin{center}
\begin{minipage}{\textwidth}\centering
\captionof{table}{Per-group AUROC for ArcFace R50 on RFW.}
\label{tab:appendix-auroc-rfw-arcface}
{%
\footnotesize
\renewcommand{\arraystretch}{1.2}
\setlength{\tabcolsep}{3.5pt}
%
%
}
\end{minipage}
\end{center}

\begin{center}
\begin{minipage}{\textwidth}\centering
\captionof{table}{Per-group AUROC for FaceNet on RFW.}
\label{tab:appendix-auroc-rfw-facenet}
{%
\footnotesize
\renewcommand{\arraystretch}{1.2}
\setlength{\tabcolsep}{3.5pt}
%
%
}
\end{minipage}
\end{center}

\begin{center}
\begin{minipage}{\textwidth}\centering
\captionof{table}{Per-group AUROC for ArcFace R100 on BFW.}
\label{tab:appendix-auroc-bfw-arcface_r100}
\resizebox{\textwidth}{!}{%
\footnotesize
\renewcommand{\arraystretch}{1.2}
\setlength{\tabcolsep}{3.5pt}
%
%
}
\end{minipage}
\end{center}

\begin{center}
\begin{minipage}{\textwidth}\centering
\captionof{table}{Per-group AUROC for AdaFace on BFW.}
\label{tab:appendix-auroc-bfw-adaface}
\resizebox{\textwidth}{!}{%
\footnotesize
\renewcommand{\arraystretch}{1.2}
\setlength{\tabcolsep}{3.5pt}
%
%
}
\end{minipage}
\end{center}

\begin{center}
\begin{minipage}{\textwidth}\centering
\captionof{table}{Per-group AUROC for MagFace on BFW.}
\label{tab:appendix-auroc-bfw-magface}
\resizebox{\textwidth}{!}{%
\footnotesize
\renewcommand{\arraystretch}{1.2}
\setlength{\tabcolsep}{3.5pt}
%
%
}
\end{minipage}
\end{center}

\begin{center}
\begin{minipage}{\textwidth}\centering
\captionof{table}{Per-group AUROC for ArcFace R50 on BFW.}
\label{tab:appendix-auroc-bfw-arcface}
\resizebox{\textwidth}{!}{%
\footnotesize
\renewcommand{\arraystretch}{1.2}
\setlength{\tabcolsep}{3.5pt}
%
%
}
\end{minipage}
\end{center}

\begin{center}
\begin{minipage}{\textwidth}\centering
\captionof{table}{Per-group AUROC for FaceNet on BFW.}
\label{tab:appendix-auroc-bfw-facenet}
\resizebox{\textwidth}{!}{%
\footnotesize
\renewcommand{\arraystretch}{1.2}
\setlength{\tabcolsep}{3.5pt}
%
%
}
\end{minipage}
\end{center}

\begin{center}
\begin{minipage}{\textwidth}\centering
\captionof{table}{Per-group AUROC for ArcFace R100 on LFW.}
\label{tab:appendix-auroc-lfw-arcface_r100}
\resizebox{\textwidth}{!}{%
\footnotesize
\renewcommand{\arraystretch}{1.2}
\setlength{\tabcolsep}{3.5pt}
%
%
}
\end{minipage}
\end{center}

\begin{center}
\begin{minipage}{\textwidth}\centering
\captionof{table}{Per-group AUROC for AdaFace on LFW.}
\label{tab:appendix-auroc-lfw-adaface}
\resizebox{\textwidth}{!}{%
\footnotesize
\renewcommand{\arraystretch}{1.2}
\setlength{\tabcolsep}{3.5pt}
%
%
}
\end{minipage}
\end{center}

\begin{center}
\begin{minipage}{\textwidth}\centering
\captionof{table}{Per-group AUROC for MagFace on LFW.}
\label{tab:appendix-auroc-lfw-magface}
\resizebox{\textwidth}{!}{%
\footnotesize
\renewcommand{\arraystretch}{1.2}
\setlength{\tabcolsep}{3.5pt}
%
%
}
\end{minipage}
\end{center}

\begin{center}
\begin{minipage}{\textwidth}\centering
\captionof{table}{Per-group AUROC for ArcFace R50 on LFW.}
\label{tab:appendix-auroc-lfw-arcface}
\resizebox{\textwidth}{!}{%
\footnotesize
\renewcommand{\arraystretch}{1.2}
\setlength{\tabcolsep}{3.5pt}
%
%
}
\end{minipage}
\end{center}

\begin{center}
\begin{minipage}{\textwidth}\centering
\captionof{table}{Per-group AUROC for ArcFace (FairFace) on LFW.}
\label{tab:appendix-auroc-lfw-arcface_fairface}
\resizebox{\textwidth}{!}{%
\footnotesize
\renewcommand{\arraystretch}{1.2}
\setlength{\tabcolsep}{3.5pt}
%
%
}
\end{minipage}
\end{center}

\begin{center}
\begin{minipage}{\textwidth}\centering
\captionof{table}{Per-group AUROC for FaceNet on LFW.}
\label{tab:appendix-auroc-lfw-facenet}
\resizebox{\textwidth}{!}{%
\footnotesize
\renewcommand{\arraystretch}{1.2}
\setlength{\tabcolsep}{3.5pt}
%
%
}
\end{minipage}
\end{center}

\subsubsection{Per-group partial AUC at low FPR}
\begin{center}
\begin{minipage}{\textwidth}\centering
\captionof{table}{Per-group Partial AUC (FPR$\le$10\%) for ArcFace R100 on RFW.}
\label{tab:appendix-pauc-rfw-arcface_r100}
{%
\footnotesize
\renewcommand{\arraystretch}{1.2}
\setlength{\tabcolsep}{3.5pt}
%
%
}
\end{minipage}
\end{center}

\begin{center}
\begin{minipage}{\textwidth}\centering
\captionof{table}{Per-group Partial AUC (FPR$\le$10\%) for AdaFace on RFW.}
\label{tab:appendix-pauc-rfw-adaface}
{%
\footnotesize
\renewcommand{\arraystretch}{1.2}
\setlength{\tabcolsep}{3.5pt}
%
%
}
\end{minipage}
\end{center}

\begin{center}
\begin{minipage}{\textwidth}\centering
\captionof{table}{Per-group Partial AUC (FPR$\le$10\%) for MagFace on RFW.}
\label{tab:appendix-pauc-rfw-magface}
{%
\footnotesize
\renewcommand{\arraystretch}{1.2}
\setlength{\tabcolsep}{3.5pt}
%
%
}
\end{minipage}
\end{center}

\begin{center}
\begin{minipage}{\textwidth}\centering
\captionof{table}{Per-group Partial AUC (FPR$\le$10\%) for ArcFace R50 on RFW.}
\label{tab:appendix-pauc-rfw-arcface}
{%
\footnotesize
\renewcommand{\arraystretch}{1.2}
\setlength{\tabcolsep}{3.5pt}
%
%
}
\end{minipage}
\end{center}

\begin{center}
\begin{minipage}{\textwidth}\centering
\captionof{table}{Per-group Partial AUC (FPR$\le$10\%) for FaceNet on RFW.}
\label{tab:appendix-pauc-rfw-facenet}
{%
\footnotesize
\renewcommand{\arraystretch}{1.2}
\setlength{\tabcolsep}{3.5pt}
%
%
}
\end{minipage}
\end{center}

\begin{center}
\begin{minipage}{\textwidth}\centering
\captionof{table}{Per-group Partial AUC (FPR$\le$10\%) for ArcFace R100 on BFW.}
\label{tab:appendix-pauc-bfw-arcface_r100}
\resizebox{\textwidth}{!}{%
\footnotesize
\renewcommand{\arraystretch}{1.2}
\setlength{\tabcolsep}{3.5pt}
%
%
}
\end{minipage}
\end{center}

\begin{center}
\begin{minipage}{\textwidth}\centering
\captionof{table}{Per-group Partial AUC (FPR$\le$10\%) for AdaFace on BFW.}
\label{tab:appendix-pauc-bfw-adaface}
\resizebox{\textwidth}{!}{%
\footnotesize
\renewcommand{\arraystretch}{1.2}
\setlength{\tabcolsep}{3.5pt}
%
%
}
\end{minipage}
\end{center}

\begin{center}
\begin{minipage}{\textwidth}\centering
\captionof{table}{Per-group Partial AUC (FPR$\le$10\%) for MagFace on BFW.}
\label{tab:appendix-pauc-bfw-magface}
\resizebox{\textwidth}{!}{%
\footnotesize
\renewcommand{\arraystretch}{1.2}
\setlength{\tabcolsep}{3.5pt}
%
%
}
\end{minipage}
\end{center}

\begin{center}
\begin{minipage}{\textwidth}\centering
\captionof{table}{Per-group Partial AUC (FPR$\le$10\%) for ArcFace R50 on BFW.}
\label{tab:appendix-pauc-bfw-arcface}
\resizebox{\textwidth}{!}{%
\footnotesize
\renewcommand{\arraystretch}{1.2}
\setlength{\tabcolsep}{3.5pt}
%
%
}
\end{minipage}
\end{center}

\begin{center}
\begin{minipage}{\textwidth}\centering
\captionof{table}{Per-group Partial AUC (FPR$\le$10\%) for FaceNet on BFW.}
\label{tab:appendix-pauc-bfw-facenet}
\resizebox{\textwidth}{!}{%
\footnotesize
\renewcommand{\arraystretch}{1.2}
\setlength{\tabcolsep}{3.5pt}
%
%
}
\end{minipage}
\end{center}

\begin{center}
\begin{minipage}{\textwidth}\centering
\captionof{table}{Per-group Partial AUC (FPR$\le$10\%) for ArcFace R100 on LFW.}
\label{tab:appendix-pauc-lfw-arcface_r100}
\resizebox{\textwidth}{!}{%
\footnotesize
\renewcommand{\arraystretch}{1.2}
\setlength{\tabcolsep}{3.5pt}
%
%
}
\end{minipage}
\end{center}

\begin{center}
\begin{minipage}{\textwidth}\centering
\captionof{table}{Per-group Partial AUC (FPR$\le$10\%) for AdaFace on LFW.}
\label{tab:appendix-pauc-lfw-adaface}
\resizebox{\textwidth}{!}{%
\footnotesize
\renewcommand{\arraystretch}{1.2}
\setlength{\tabcolsep}{3.5pt}
%
%
}
\end{minipage}
\end{center}

\begin{center}
\begin{minipage}{\textwidth}\centering
\captionof{table}{Per-group Partial AUC (FPR$\le$10\%) for MagFace on LFW.}
\label{tab:appendix-pauc-lfw-magface}
\resizebox{\textwidth}{!}{%
\footnotesize
\renewcommand{\arraystretch}{1.2}
\setlength{\tabcolsep}{3.5pt}
%
%
}
\end{minipage}
\end{center}

\begin{center}
\begin{minipage}{\textwidth}\centering
\captionof{table}{Per-group Partial AUC (FPR$\le$10\%) for ArcFace R50 on LFW.}
\label{tab:appendix-pauc-lfw-arcface}
\resizebox{\textwidth}{!}{%
\footnotesize
\renewcommand{\arraystretch}{1.2}
\setlength{\tabcolsep}{3.5pt}
%
%
}
\end{minipage}
\end{center}

\begin{center}
\begin{minipage}{\textwidth}\centering
\captionof{table}{Per-group Partial AUC (FPR$\le$10\%) for ArcFace (FairFace) on LFW.}
\label{tab:appendix-pauc-lfw-arcface_fairface}
\resizebox{\textwidth}{!}{%
\footnotesize
\renewcommand{\arraystretch}{1.2}
\setlength{\tabcolsep}{3.5pt}
%
%
}
\end{minipage}
\end{center}

\begin{center}
\begin{minipage}{\textwidth}\centering
\captionof{table}{Per-group Partial AUC (FPR$\le$10\%) for FaceNet on LFW.}
\label{tab:appendix-pauc-lfw-facenet}
\resizebox{\textwidth}{!}{%
\footnotesize
\renewcommand{\arraystretch}{1.2}
\setlength{\tabcolsep}{3.5pt}
%
%
}
\end{minipage}
\end{center}

\end{document}